\documentclass[letterpaper]{article} %
\usepackage{aaai2027}  %
\usepackage{rotating}

\usepackage[hyphens]{url}  %
\usepackage{graphicx} %

\usepackage{natbib}  %
\usepackage{caption} %
\usepackage{algorithm}
\usepackage{algorithmic}
\usepackage{amsmath}
\usepackage{amsfonts}
\usepackage{amssymb}
\usepackage{amsthm}

\usepackage{newfloat}
\usepackage{listings}
\DeclareCaptionStyle{ruled}{labelfont=normalfont,labelsep=colon,strut=off} %
\floatstyle{ruled}
\newfloat{listing}{tb}{lst}{}
\floatname{listing}{Listing}

\usepackage{booktabs}

\nocopyright %

\title{Learning through Capacity Awareness to Minimize Hospital Readmissions}

\author {
    Lydia T. Liu\textsuperscript{\rm 1}\equalcontrib,
    Kara Schechtman\textsuperscript{\rm 1}\equalcontrib,
    Pengyi Shi\textsuperscript{\rm 2}\equalcontrib
    , Angela Zhou\textsuperscript{\rm 3}\equalcontrib
}
\affiliations {
    \textsuperscript{\rm 1}Princeton University\\
    \textsuperscript{\rm 2}Purdue University\\
    \textsuperscript{\rm 3}University of Southern California\\
    ltliu@princeton.edu, schechtk@princeton.edu, shi178@purdue.edu,zhoua@marshall.usc.edu
}
\usepackage{mathtools}
\DeclarePairedDelimiter\abs{\lvert}{\rvert}

\newcommand{\norm}[1]{\left\|{#1}\right\|}

\renewcommand{\Pr}{\mathbb{P}}

\newtheorem{example}{Example}[section]

\newtheorem{theorem}{Theorem}[section]
\newtheorem{definition}{Definition}[section]
\newtheorem{proposition}[theorem]{Proposition}
\newtheorem{corollary}[theorem]{Corollary}
\theoremstyle{definition}

\newtheorem{fact}[theorem]{Fact}

\begin{document}
\pagestyle{plain} %

\maketitle

\begin{abstract}
	
Predictive algorithms are increasingly used to guide hospital discharge decisions, yet these models can exhibit censoring bias for vulnerable patients due to insufficient historical data or past discharge decisions. 
	When biased readmission risk predictors are used to inform discharges, the training data generated by discharge decisions may perpetuate similar biases when predictors are re-trained, leading to a vicious cycle.
	We extend a queueing-based Markov decision process for hospital discharge to incorporate periodically retrained readmission risk predictors. Our key insight is that cost-optimal, ``capacity-aware'' operational decision policies can naturally diversify the lengths of stay at which patient outcomes are observed, enabling consistent estimation of readmission risk without explicit fairness constraints.
	We prove that repeated retraining under discharge-up-to policies converges to unbiased risk estimates under well-specification. In synthetic experiments and calibrated simulations based on patient-level data from a partner hospital, we find that where commonly used threshold-on-risk policies can remain persistently biased, discharge-up-to policies reduce learning error over status quo decision policies (such as thresholding on risk) by 24\%--25\%. These improvements to estimation also mitigate some forms of bias, reducing readmission cost disparities (adjusted for baseline risk differences between demographic groups) by 38\%--63\%. Our results demonstrate how capacity-aware decision policies can improve both predictive performance and fairness by addressing censoring bias.

\end{abstract}

\begin{links}
    \link{Code}{https://.}
\end{links}

\section{Introduction}

Data driven algorithms learn to predict from decisions made in the past. This ``selective labels'' problem \citep{lakkaraju2017selective,rambachan2020bias} arises whenever outcomes are observed only for cases that receive a particular decision: loan repayment only for approved applicants, recidivism only for released defendants \citep{kleinberg2018human}. This selectivity can entrench bias when models are retrained on the resulting censored data \citep{Ensign18,hashimoto18demographics,liu2020disparate,kallus2018residual}. %
\citet{Kilbertus19Fair} model such settings with \emph{static threshold} decision rules with binary action space, where the action (e.g., approve, release) is optimal whenever a predicted risk crosses a fixed threshold. They show that introducing random
\emph{exploration}, making random decisions that deviate from the optimal policy, is key to learning from biased on-policy data.

We revisit this problem in a setting where selective observation arises not from a static threshold, but from a \textbf{dynamic, capacity-constrained decision}: hospital discharge decisions informed by readmission risk \citep{shi2021timing}. Here, cost and benefit of discharge depends on system-level state (e.g., hospital ward congestion) as well as individual risk, so the mapping from risk to discharge is random and state-contingent. Moreover, arbitrary exploration is ethically impermissible for hospital discharge decisions. This raises a fundamental question: When do dynamic, state-dependent decision policies entrench data bias?  How can we create virtuous cycles in resource-constrained settings, such that decisions generate the data needed to correct biased predictions over time?

Such questions are pressing in healthcare, where racial and socioeconomic disparities in hospital readmission rates are well documented \citep{liu2020reporting,gichoya2021equity,tsai2014disparities,rambachan2022racial,lizza2021racial,monangi2022racial}. Broad interventions like Medicare's Hospital Readmission Reduction Program have in some cases worsened disparities \citep{joynt2019adjusting,chaiyachati2018changes}. While predictive risk models can be used to identify high risk patients and inform clinicians \cite{napatientoutcome}, they inherit \emph{censoring bias} from institutional decisions that generate the training data: if existing practice generates a lack of data on the benefits of extended stays for underserved patients, models can misidentify risk and perpetuate disparities in care and health outcomes.

In this paper, we find that a class of capacity-aware discharge policies, known as \emph{discharge-up-to} policies, result in recoverable distribution shift, in contrast to static decision systems that induce unrecoverable distribution shift. Under natural variation in patient arrivals, discharge-up-to policies generate sufficiently diverse observations of patient outcomes with full-support over patient and length-of-stay covariate distributions, without introducing suboptimal randomized decisions or explicit fairness constraints. In contrast, censoring bias in static decision systems results in never-observed covariate regions that require additional explicit exploration \citep{Kilbertus19Fair}.

Our contributions are as follows.
\textbf{(1)} We formulate repeated hospital discharge as a dynamic selective-labels problem, extending beyond settings with known readmission risk to settings where risk must be learned from policy-generated, potentially biased data (\S\ref{sec:stylize-model}).
\textbf{(2)} We prove that, under stochastic arrivals, discharge-up-to policies discharge every patient type at every length of stay with positive probability, and that the full support of resulting observations is sufficient to recover the true risk curve via retraining when prediction models are well-specified (\S\ref{sec:theory:discharge_policies}--\ref{sec:theory}).
\textbf{(3)} We empirically validate our findings in simulations against fixed-threshold and myopic dynamic threshold baselines, finding that discharge-up-to policies reduces prediction error, improves total cost, adapts better across cost regimes, reduces disparities in prediction error, and reduces readmission cost disparities (adjusted for baseline risk) (\S\ref{sec:empirical}). %

\section{Background and Related Work}\label{sec:background}

\paragraph{Censoring bias in readmission prediction.}
Common predictive models of patient severity and readmission risk suffer from bias that can perpetuate or even exacerbate disparities in downstream decision-making. \citet{wang2022bias} identify measurement bias as a driver of inequitable readmission prediction in four widely used models. Measurement bias arises when the quality, quantity, and variety of collected data differ systematically across demographic groups. This includes what we call \emph{censoring bias}, which occurs when historical decisions determine which outcomes are observed, leaving outcomes under alternative decisions (e.g. longer stays) for some populations systematically omitted from available data.

\paragraph{Feedback loops.}
When biased predictions drive decisions that further skew the training data, a vicious cycle can result. \citet{obermeyer2019dissecting} document a prominent example: a widely deployed algorithm predicted future healthcare costs as a proxy for severity. Because Black patients historically incurred lower costs due to reduced access to care, the model systematically underestimated their needs and downstream decisions denied Black patients care, compounding the bias. More broadly, \citet{Ensign18} and \citet{rambachan2020bias} study how feedback loops entrench biased decisions in sequential settings, and \citet{yan2022observability} show that when a demographic variable affects both data observability and patient outcomes, bias remediation is complex. 

Whether biased data produce self-censoring decisions (the so-called vicious cycle of ``bias-in, bias-out'') depends on how decisions affect individual welfare and the observed labels distribution \citep{rambachan2020bias,kallus2018residual}. Prior analyses study binary decisions where selection has constant individual costs and static decision thresholds map deterministically to censored distributions. By contrast, in hospital discharge, costs depend on patients' length of stay and hospital congestion, and data censoring patterns of decision policies can vary with system state.

In static settings, \citet{Kilbertus19Fair} show that exploratory decisions are necessary to learn from censored data and propose an $\epsilon$-greedy decision policy. Our work shows that capacity-aware discharge-up-to policies provide a patient-centered alternative, by extending stays when beds are available for groups that would otherwise be censored. Natural exploration enables the prediction model to learn without randomized experimentation.

\paragraph{Capacity-aware discharge policies.}
\citet{shi2021timing} introduce the discharge-up-to policy for balancing readmission risk against inpatient congestion in a queueing model and prove its optimality under structural conditions on the risk function. However, their analysis assumes the readmission risk model is known and correct. In practice, risk models are estimated from observational data censored by past decisions, with potential differences in censoring bias across groups. Our work thus asks: how do discharge-up-to policies shape the \emph{learning dynamics} of the risk model, and can capacity-aware policies correct censoring bias over time?

\paragraph{Prediction and priorization under capacity constraints.}
Capacity constraints shape prediction in other high-stakes allocation settings like housing and education, where they raise tradeoffs between efficiency, equity, and learning \citep{torosprioritize,kube2019allocating,kube2023community,diaz2026predicting,wilder2025learning,moon2026paradox} and can make outcomes less predictable \cite{dong2026capacity}. Our setting differs in that capacity constraints affect not only patients' interventions or outcomes, but also observability of outcomes for different lengths of stay and therefore what future models learn.

\section{Stylized Model}\label{sec:stylize-model}

We adopt the dynamic queueing model of \citet{shi2021timing}, which formulates hospital discharge as a Markov decision process with multiple patient types and capacity constraints. The hospital must balance two competing objectives: discharging patients early increases readmission risks, while keeping patients longer increases congestion costs. Our key departure from prior work is that in our setting, true risk curves are unknown and must be estimated from data as well as updated over time via retraining. A complete notation table is in Appendix~\ref{app:notation}.

\subsection{Queueing Model}

Each day $k \in \{1, \ldots, T\}$, patients arrive according to a Poisson process with rate $\lambda > 0$. Each arrival is independently assigned to type $w \in \mathcal{W} = \{1, \ldots, M\}$ with probability $\pi_w > 0$, so the number of type-$w$ arrivals on day $k$, $A_{k,w} \sim \text{Poisson}(\lambda\pi_w)$. Patients remain in the hospital for a \emph{length of stay} (LOS) $\ell \in \{1, \ldots, J\}$, where $J$ is the maximum allowed LOS; patients reaching LOS $J$ are automatically discharged. Unlike conventional queueing models with i.i.d.\ service times, service times (LOS) here are \textbf{policy-dependent}: the hospital's discharge policy determines each patient's actual LOS.
Each day $k$, the hospital observes the \textbf{system state} $X_k=(X_{k,w}^\ell)_{w \in \mathcal{W}, \ell \in [J]} \in \mathbb{N}^{M \times J}$ (occupancy counts of each type at each LOS). A \textbf{discharge policy} $\Pi(X_k)$ then produces a discharge decision $D_k = (D_{k,w}^\ell)_{w \in W,\, \ell \in [J]}$, where $D_{k,w}^\ell$ is the number of type-$w$ patients at length of stay $\ell$ to discharge on day $k$. We discuss different discharge policies in \S\ref{sec:theory:discharge_policies}.

Undischarged patients advance to LOS $\ell+1$ the next day and new arrivals $A_{k,w} $ enter the system and advance to LOS $1$. The type $w$ \textbf{patient census} evolves as $S_{k+1,w} = S_{k,w} - \|D_{k,w}\|_1 + A_{k,w}$, where $\|\cdot\|_1$ denotes the $L_1$ norm (i.e., total number of discharges on day~$k$). We model an infinite occupancy queue for tractability, and capture capacity constraints through the congestion cost instead (\S\ref{sec:decision_costs}).

\subsection{Readmission Risk}\label{sec:erm}

For each type $w$ and LOS $\ell$, the \emph{readmission risk} $r_w(\ell)$ represents the probability that a type $w$ patient discharged at LOS $\ell$ will be readmitted within a fixed time window (e.g., 90 days):
\begin{equation*}\label{eq:risk_function}
r_w(\ell) = \mathbb{P}(\text{readmission} \mid \text{type } w, \text{ discharged at LOS } \ell).
\end{equation*}

The risk function can depend on the patient type $w$, capturing the fact that different patient types may have different risk profiles: $r_{w}(\ell) \neq r_{w'}(\ell)$ for $w \neq w'$.\footnote{More generally, one can model risk jointly as $r(z)$ where $z = (\ell, w)$ collects all covariates: $w$ captures fixed patient characteristics at admission, while $\ell$ is the operational decision the hospital controls. The distribution of $z$ in the training data depends on the discharge policy, a point we return to in Section~\ref{sec:theory}.} For example, patients from disadvantaged communities may have higher baseline readmission risk at all LOS values due to factors such as social determinants of health, access to post-discharge care, and chronic conditions.

\paragraph{Risk Prediction via Empirical Risk Minimization.}
In practice, the hospital does not observe the true risk function $r_w(\ell)$ but instead estimates it from historical data. We denote the estimated risk predictor for type $w$ as $\hat{r}_w(\ell)$. These predictors are learned via empirical risk minimization (ERM) on observed readmission outcomes from previously discharged patients.
At regular intervals (e.g., monthly), the predictor is retrained on accumulated discharge data. We denote the dataset available for retraining round $t$ as
$
\mathcal{D}_t := \{(w_i, \ell_i, Y_i) : \text{ patient $i$ discharged before round } t\}
$
where $w_i$ is patient type, $\ell_i$ is observed LOS, and $Y_i \in \{0,1\}$ is the readmission outcome.

At round $t$, the predictor $\hat{r}_{w,t}(\ell)$ for type $w$ patients is updated according to:
\begin{equation}\label{eq:erm}
\hat{r}_{w,t+1}
= \arg\min_{r \in \mathcal{R}}
\sum_{(w_i, \ell_i, Y_i) \in \mathcal{D}_t: w_i = w}
h\big(r_{w_i}( \ell_i), Y_i\big),
\end{equation}
where $h(\cdot,\cdot)$ is a prediction loss function (e.g., cross-entropy) and $\mathcal{R}$ denotes the model class (see Example~\ref{ex:logistic}).
Critically, because readmission outcomes are only observed for \emph{discharged} patients \emph{at their respective lengths of stay}, the training data is \emph{censored} by the discharge policy itself. This creates potential feedback loops where biased predictors lead to biased discharge decisions, which in turn produce biased training data for future model updates.

\begin{example}[Logistic Regression Risk Model]\label{ex:logistic}
A common choice for the model class $\mathcal{R}$ is logistic regression, where the risk function takes the form
\begin{equation}
    \ln\left( \frac{r_w(\ell)}{1-r_w(\ell)}\right) = b_{1,w} \cdot \ell + b_{0,w}.
\label{eq:logit-risk}
\end{equation}
Here, for patient type $w$, $b_{0,w} > 0$ is the intercept term (baseline risk value), and $b_{1,w} < 0$ is the \emph{treatment effect} of the length-of-stay. The condition $b_{1,w} < 0$ captures the clinical intuition that extending LOS reduces the risk of readmission, as patients benefit from additional treatment and observation. Both the baseline and the treatment effect can depend on the patient type $w$.
\end{example}

\subsection{Hospital Decision Costs}\label{sec:decision_costs}
Following~\citet{shi2021timing}, the hospital incurs two costs each day $k$: a \emph{readmission cost} $c_R(D_k) = c_R \cdot \sum_{w,\ell} D_{k,w}^\ell \, r_w(\ell)$ penalizing discharges proportional to their readmission risk, and a \emph{congestion cost} $c_h(S_k, D_k)$. The congestion cost captures operational capacity constraints in hospitals: keeping patients longer increases staffing demands, resource usage, and may reduce quality of care. Suitable choices of congestion cost function $c_h$ and scale parameter $h$ capture the cost's dependence on the overall patient census $S_k$, for instance the quadratic congestion cost  $c_h(S_k, D_k) = h \cdot (S_k - \|D_k\|_1)^2$ or overage cost above a fixed capacity limit $N$, $c_h(S_k,D_k)=h\cdot \max(0,S_k-N).$ The hospital seeks to minimize the long-run average total cost, which sums readmission and congestion costs:
\begin{equation}\label{eq:long-run-cost}
\lim_{T \to \infty} \frac{1}{T} \sum_{k=1}^T \big( c_R(D_k) + c_h(S_k, D_k) \big).
\end{equation}
The ratio $c_R / h$ governs the key decision tradeoff faced by the hospital: high $c_R$ favors retaining patients to reduce readmissions; high $h$ favors discharging to relieve congestion.

\section{Discharge Policies}\label{sec:theory:discharge_policies}

In this section, we describe the discharge-up-to policy for discharge decisions and establish its optimality properties. We describe other policy classes, including comparison threshold-based policies, in Appendix~\ref{app:threshold-policies}.

\subsection{Discharge-Up-To Policy}\label{def:discharge_up_to_policy}

Our main policy class is the \emph{discharge-up-to} policy class, $\Pi_{\tilde{D}}$. The key idea is to target a census level: on each day $k$, patients are discharged in a priority order until the number of patients remaining in the hospital equals the target level. This separates the discharge decision into two parts: a target \emph{discharge-up-to level}  $\tilde{D}_k$, which determines how many patients should remain after discharge decisions, and a priority ordering $\succ$, which determines which patients are discharged first when the hospital is above that target.

Let $\succ$ denote such a priority ordering over patient-type and length-of-stay pairs $(w, \ell)$ for $w \in \mathcal{W}$, $\ell \in \{1, \ldots, J-1\}$ (see below). 
We define
\begin{equation}
\Lambda_k^{w,\ell} = \sum_{\substack{w' \in W,\, j \leq J \\ (w',j) \succ (w,\ell)}} X_{k,w'}^j 
\end{equation}
as the number of patients who will already have been discharged or automatically leave before the policy reaches type-$w$ patients at LOS $\ell$, that is, those with higher priority for discharge $(w', j) \succ (w, \ell)$, including patients at the maximum LOS $J$. The number of patients remaining when the policy reaches $(w, \ell)$ is therefore $S_k - \Lambda_k^{w,\ell}$, and the policy discharges from this group only as many as needed to reach the target $\tilde{D}_k$:
\begin{equation}\label{eq:discharge_up_to_decision}
D_{k,w}^\ell = \min\Bigl(X_{k,w}^\ell,\; \max\bigl(S_k - \Lambda_k^{w,\ell} - \tilde{D}_k,\; 0\bigr)\Bigr).
\end{equation}
That is, up to $X_{k,w}^\ell$ patients are discharged from group $(w,\ell)$, but only to the extent needed to bring the census down to $\tilde{D}_k$; if the target has already been reached by higher-priority discharges, no patients from $(w, \ell)$ are discharged. Ties in priority are broken by random sampling among tied patients. Formally, the discharge-up-to policy is $\Pi_{\tilde{D}}(X_k) := \bigl(D_{k,w}^\ell\bigr)_{w \in W,\, \ell \in [J]}$, where $D_{k,w}^\ell$ is given by \eqref{eq:discharge_up_to_decision} and $D_{k,w}^J := X_{k,w}^J$.

\subsubsection{Optimality of the Discharge-Up-To Policy}\label{sec:optimality}

The discharge-up-to policy is important because it is known to be optimal for the cost-minimization problem under a structural condition on the priority ordering. Specifically, \citet[][Corollary~1]{shi2021timing} establish that, whenever the patient priority ranking satisfies \emph{strong dominance} (see Appendix~\ref{app:dominance}), the optimal policy under \emph{known readmission risks} for minimizing the long-run average cost~\eqref{eq:long-run-cost} takes the form of a discharge-up-to policy. That is, the optimal policy discharges patients by the strong dominance priority order until the remaining census reaches a state-dependent target level $\tilde{D}^*(X_k)$.

Strong dominance requires that the priority ordering rank patients by \textit{\textbf{marginal benefits}} from additional hospitalization $r(\ell) - r(\ell+k)$ in a uniform sense: if $(w,\ell)$ is prioritized for discharge ahead of $(w',\ell')$, then extending the stay of a type-$w$ patient from LOS $\ell$ must result in a weakly smaller risk reduction than extending the stay of a type-$w'$ patient from LOS $\ell'$, for every feasible extension length $k$. This is a significant point of departure from status quo approaches (Appendix \ref{app:threshold-policies}), which discharge patients by thresholding on absolute readmission risk. In general, there need not exist a ranking of patients that satisfies strong dominance. For a single patient type, a sufficient condition for the existence of a strong dominance ranking is diminishing marginal returns of hospitalization, i.e., $r(\ell) - r(\ell+1)$ is non-increasing in $\ell$ (Appendix~\ref{app:dominance} Fact~\ref{fact:strong_dom_M1}). Existence conditions for $M \geq 2$ types are given in Appendix~\ref{app:dominance}. In practice, the notion of \emph{weak dominance} introduced in \citet{shi2021timing} (Appendix~\ref{app:weak_priority_order}) allows one to construct priority orders from estimated risk curves, for which a priority order satisfying strong dominance may not exist.

In Proposition~\ref{prop:optimal} (Appendix~\ref{app:discharge-up-to}), we provide a \textbf{complementary optimality result} for discharge-up-to policies that relaxes the strong dominance assumption. Using a steady-state approximation of the LOS distribution and a quadratic approximation of the value function, we show that the optimal discharge policy takes the discharge-up-to form with target level $\tilde{D}_k$, where
\begin{itemize}
    \item \textbf{Higher readmission cost $c_R$} $\Rightarrow$ \textbf{higher optimal target} $\tilde{D}_k$ (fewer discharges, longer stays to reduce readmits);
    \item \textbf{Higher congestion cost $h$} $\Rightarrow$ \textbf{lower optimal target} $\tilde{D}_k$ (more discharges to relieve capacity pressure).
\end{itemize}
In practice, grid search can be used to approximate the optimal discharge-up-to level. Both optimality results  (Proposition~\ref{prop:optimal} and Cor. 1 from \citet{shi2021timing}) assume the risk model is \emph{known and fixed}. In practice,  readmission risk must be learned from data generated by the discharge policy itself (\S\ref{sec:erm}). In the next section, we analyze how the discharge-up-to policy's capacity-driven decisions shape the learning dynamics of the risk model.

\section{Theoretical Analysis}\label{sec:theory}

Our main result shows that under stochastic arrivals, discharge-up-to policies do not censor outcomes for any patient type: every possible length of stay occurs with positive probability. Consequently, assuming that the true risk model is in the parametric family of the estimator, ERM \eqref{eq:erm} is asymptotically consistent. That is, patient readmission data collected under discharge-up-to policies eventually enable successful estimation of the readmission risk model. %
These convergence results do not depend on the initial risk predictor used for decision making. In other words, under the natural randomness of arrivals that is common in hospitals and other service systems, \textit{learning under discharge-up-to policies is asymptotically robust to any initially biased risk predictor.}

\begin{figure*}[ht]\centering
\includegraphics[width=0.9\textwidth]{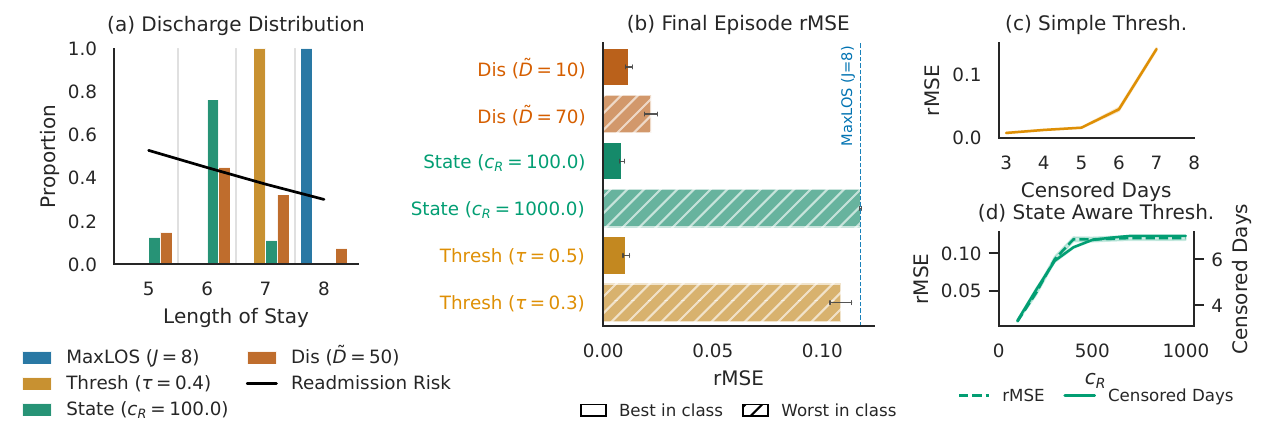}
\caption{(a) Distribution of discharge decisions made by the discharge-up-to policy and baselines over 100,000 days. Policies decide discharges using a univariate logistic model on LOS with $\hat{r}(1)=0.7$, $\hat{r}(8)=0.3$. (b) $rMSE$ of the best and worst-performing parameter in each policy class at the final episode for a representative risk model configuration ($r(1)=\hat{r}_1(1)=0.7$, $r(8)=\hat{r}_1(8)=0.3$). (c) Number of censored days in the final episode vs. the simple threshold policy's $rMSE$. (d) Readmission cost vs. state aware policy's $rMSE$ and number of censored days in the final episode.}\label{fig:rmse}
\end{figure*}

To build intuition, let us first consider simulation evidence in a setting where the patient risk model is known and fixed.
Figure~\ref{fig:rmse}a illustrates how different discharge policies affect the distribution of observed LOS (LOS) under a hypothetical univariate logistic risk model. As expected, the Max LOS policy censors readmission outcomes for all lengths of stay below the maximum.
Policies based on fixed or state-aware risk thresholds (defined in Appendix~\ref{app:threshold-policies}) concentrate discharges within a narrow LOS range, censoring outcomes of shorter or longer stays.
In contrast, the discharge-up-to policy spreads discharges over a broader range of LOS values, resulting in a more diverse set of observed outcomes, because the policy implicitly introduces stochasticity through fluctuating occupancy and arrival levels.

Stochastic arrivals are necessary to prevent self-perpetuating censoring cycles. We give a counterexample of \textit{bias-in, bias-out} dynamics where given a biased initialization and for a range of discharge-up-levels, \textit{no} data is generated for a particular group at a range of LOS, perpetuating bias. We give an informal presentation of the main idea for brevity; see Appendix \ref{app:bias-in-bias-out} for the formal result and proof. 

\begin{proposition}[Distribution support failure under deterministic arrivals
(informal)]
Suppose arrivals are constant per group ($n_a, n_b$), and the priority order induced by the
current learned predictor has group-specific cutoff LOS's
\(L_b<L_a\). For every target $\widetilde D$ satisfying
\[
n_a(L_a-1)+n_b(L_b-1)
\le \widetilde D
\le n_aL_a+n_b(L_b-1),
\]
the corresponding discharge-up-to policy has an operational steady
state in which type \(b\) is discharged at \(L_b\) before \(L_a\)

The steady-state distribution has a region of $0$ support, $d^\Pi(b,\ell)=0, \ell>L_b,$ so the risks \(r_b(\ell)\) for \(\ell>L_b\) are not identified
and asymptotic consistency is not possible.
\end{proposition}
In particular,
if the true-risk oracle strictly prefers to retain type \(b\) beyond
\(L_b\), then this operational steady state is strictly suboptimal, and cannot generate data to correct it.

We formalize the non-censorship of discharge-up-to-policies under stochastic arrivals in the following theorem.

\begin{theorem} [Non-censorship under discharge-up-to policies and stochastic arrivals.]
\label{thm:main}
Fix a maximum length of stay $J \ge 2$ and a discharge-up-to policy with target level $c>0$. Let exogenous arrivals $\{A_k\}_{k\ge 1}$ be i.i.d.\ with
$\Pr(A_k=a)>0$ for all integers $a\ge 0$.\footnote{In particular, this holds for $A_k\sim\mathrm{Poisson}(\lambda)$ with $\lambda>0$.}
Each arrival is independently assigned one of $M$ types, where type $w$ has population proportion $\pi_w > 0$. Denote arrivals of type $w$ on day $k$ as $A_{k,w}$.
Assume
\begin{enumerate}
    \item patients cannot be discharged on the day they arrive;
\item all patients are discharged when they reach day $J$; and \item \textbf{weak preference for new arrivals}: when any discharges occur, patients of type $w^*$ with the minimum $\textrm{LOS} = 1$ are (weakly) the lowest priority to be discharged, that is, \[ \forall w, \ell, ~(w, \ell) \succeq (w^*, 1)\] and that all ties are brokenly randomly.\footnote{Note that this is satisfied by random discharge priority as well as any priorization scheme that meets the strong dominance criterion.}
\end{enumerate}

Then, for any fixed patient of any type $w$ who enters the system, their length of stay $L$
satisfies
\[
\Pr(L=j) > 0 \qquad \text{for every } j\in\{1,2,\dots,J\}.
\]
\end{theorem}
\begin{proof}[Proof sketch (full proof in Appendix~\ref{app:non-censorship-proof})]
Fix an arbitrary patient $i$ arriving on day $k$. The proof proceeds by induction on the length of stay $j$, establishing that for each $j \in \{1, \dots, J\}$, both $\Pr(L = j) > 0$ and $\Pr(L \geq j+1) > 0$. The two key observations are:
\begin{enumerate}
\item \emph{Discharge is possible at any LOS.} If a sufficiently large burst of arrivals occurs (specifically, $A_{k+j-1,w^*} = c$), the census exceeds the target $c$ and the policy must discharge patients, including, with positive probability under tie-breaking, a patient at LOS $j$. Arrival rate assumption guarantees such bursts occur with positive probability.
\item \emph{Survival to the next LOS is also possible.} If zero arrivals occur on a given day ($A_{k+j-1} = 0$), the post-discharge census remains at or below $c$, so no further discharges are needed and the patient advances to $\mathrm{LOS} = j+1$.
\end{enumerate}
Since both events occur with positive probability, the induction carries through to $j = J$, completing the argument.
\end{proof}

Although there is positive probability of viewing every length of stay in the discharged data, the distribution of observed discharged patients need not match the distribution of the same under the optimal policy. This creates a ``covariate-shift'' problem:  while the underlying readmission-risk function to be learned is fixed, the distribution of prediction covariates (LOS and patient type) changes.

Standard tools from the covariate shift literature can be used to adjust the readmission-risk learning problem for this distribution shift. We let $z_t := (\ell_t, w_t)$ denote the individual patient-level information that readmission risk depends on.
For a fixed discharge policy $\Pi$, we let
$
d^{\Pi}(z) := \lim_{t \to \infty} \Pr_{\Pi}(z_t = z), z = (\ell,w),$
denote the stationary distribution of patient-level information at the time of potential discharge. Since patient types are determined by exogenous arrivals, distribution shift primarily arises through the length-of-stay distribution, \(\Pr_{\Pi}(\ell_t\mid w_t)\), which is policy-dependent.

The implications of this distribution shift, formalized in two corollaries below, depend sharply on whether the risk model is well specified. Under correct specification, ERM can recover the same underlying risk function despite covariate shift. Under misspecification, as is common in practice, the limiting ERM solution may depend on the policy-induced covariate distribution. To highlight this distinction, we let the risk model $r_\theta$ be parametrized by $\theta \in \Theta$ where $\Theta \subseteq \mathbb{R}^d$ for some finite $d$. Denote the unweighted empirical risk minimizer on the discharged subpopulation as $\theta^1_t$:
\begin{equation}
\theta^1_t \in \arg\min_{\theta \in \Theta} E_{z_t\sim d_\Pi^t}[h\left(r_{\theta}\left(z_t\right), Y\right) \mathbb{I}[D_t = 1]]. 
\label{eq-dischargedistribution}
\end{equation}
\begin{corollary}[Unbiased Learning under well-specified prediction]\label{cor:wellspec}
Suppose the risk model is well-specified, i.e. it is estimated over a parametric family $\Theta$ that indexes the true readmission risk $\mathbb{P}(Y=1\mid \ell, W),$ and there is no model misspecification. Then unweighted empirical risk minimization is asymptotically consistent, $\hat \theta^1 \to \theta^*.$
\end{corollary}

\begin{corollary}[Inverse propensity weighted estimation (IPW)]\label{cor:ipw}
    Suppose the parametric family for $r(\ell)$ is misspecified, i.e. $\theta^1 \neq \theta^*.$ Consider the probability-of-discharge reweighted empirical risk minimizer: 
    \begin{equation}\tilde{\theta} \in \arg\min_{\theta \in \Theta} E_{z_t \sim d_\Pi^t}\left[h\left(r_{\theta}\left(z_t\right), Y\right)
   \frac{ \mathbb{I}[D_t = 1]}{\Pr(D_t=1 \mid z_t)}
    \right]. 
\label{eq-dischargedistribution2}
\end{equation}
Then $\tilde{\theta}\to \theta^*.$
\end{corollary}

These corollaries follow from standard results in covariate shift and reweighted maximum likelihood, e.g. \citep[Sec. 3]{shimodaira2000improving}. When the risk model $r_\theta$ is misspecified, unadjusted maximum likelihood estimation recovers a parameter that is the KL-divergence projection of the true function onto the parameter space, which is biased \citep{shimodaira2000improving}. Theorem~\ref{thm:main} verifies the main assumption required, namely that importance sampling weights would be strictly bounded away from zero. To address this issue, IP reweighting by the probabilities of discharge is sufficient to adjust for distribution shift that arises in different policies. In experiments (Appendix~\ref{app:simulations:extensions}), we found that IPW slightly improves performance over ERM in a simple mis-specified setting.

\section{Simulations}\label{sec:empirical}

In this section, we evaluate whether the discharge-up-to policy improves finite-sample learning and reduces costs in realistic simulations, complementing our theoretical guarantees. Motivated by the prevalence of predictive biases in readmission risk models, we investigate whether the discharge-up-to-policy can improve racial readmission disparities and correct inaccurate predictions through natural exploration.

We provide a brief overview of the experimental setup. Full simulation
details Appendix~\ref{app:simulations:methods}.

\paragraph{Simulation Setup.}
We simulate two queue settings over 50 retraining episodes of $T=60$ days, with maximum LOS $J=8$, the $95^{th}$ percentile  LOS in our dataset. In the fully synthetic \textbf{simple setting}, we use one patient group, Poisson arrivals with rate $\lambda=9.4$, and logistic true and estimated risk models with LOS as the only covariate. Testing 50 risk/bias configurations, this setting isolates how censoring and exploration affect learning. In the \textbf{realistic setting} calibrated to hospital data, arrivals follow a Poisson process with rate $\lambda=6.5$ (matching the historical arrivals rate) and patient covariates sampled from the empirical distribution. True risk is generated from the logistic cure model of \citet{shi2021timing}, with covariates $Z=(\log(\ell),Z^e)$, where $Z^e$ are policy-exogenous patient features; discharged patients readmit with probability $r(Z)=1-\sigma(\xi^{\mathsf{T}}Z+\xi_0)$. Coefficients $\xi, \xi_0$ are learned from the hospital dataset. $\hat r_1$ is constructed from $r$ to underestimate risk for African-American patients. We find similar results when risk is underestimated for Caucasian patients (Appendix \ref{app:simulations:extensions}).

\paragraph{Dataset.} The realistic data experiments use patient-level admissions data from a hospital in Indiana from January 2010 to September 2017. 
After exclusions, the dataset includes $n=17,947$ patients ($16,195$ Caucasian, $1,554$ African-American, and $198$ other). 
Because we study de-identified, archived data, the Purdue IRB determined that it does not meet the regulatory definition of human subjects research. Due to a data use agreement in place with the hospital, the dataset cannot be publicly released. To support reproducibility, we include simulations in a fully synthetic setting and will release the code in a public repository.

\paragraph{Evaluation Baselines and Metrics.} We compare the discharge-up-to policy to two baseline policies, \emph{fixed-threshold} and \emph{state-aware threshold} (more details in Appendix \ref{app:threshold-policies} and \ref{app:parameters}). In the simple setting, the discharge-up-to policy uses fixed target levels $\tilde{D}$ from 10--70, and in the realistic setting, we augment the policy with a grid search to tune the level. We highlight results with the level initialized to $\tilde{D}_1=30$; all starting levels perform similarly. We also consider a conservative \textit{maximum LOS baseline} that holds all patients $J=8$ days.

\emph{Learning.} We evaluate root mean squared error ($rMSE$) of each model iterate $\hat{r}_i$. In the simple setting, we average uniformly over LOS to make comparisons across policies with different discharge distributions. 
In the realistic setting, we compute~$rMSE$~against individual risk predictions and average error over the empirical distribution of patient covariates and the uniform distribution over possible LOS. For a particular patient group $w$, with $\mathcal{D}^w=\{z^e_j\in \mathcal{D}: w_j=w \}$ and $z=(\log(\ell),z^e)$, we report
\begin{equation*}
rMSE(i,w) =  \sqrt{ \mathbb{E}_{z^e\sim \mathcal{D}^w,\ell\sim \mathcal{U}(1,J)} [(r(z) - \hat{r}_{i}(z))^2]}.
\end{equation*}

\emph{Costs.} 
We evaluate the performance of discharge policies using total hospital cost, decomposed into congestion and readmission costs. We set $N=20$, $h=1$ and $c_R=5$; in Appendix \ref{app:simulation:results}, we present results varying $c_R$ to study how policies trade off between congestion and readmission under different cost regimes. 

\emph{Equity.} For the realistic setting, we measure disparities in learning performance using group differences in $rMSE$, i.e. for two groups $w_1$ and $w_2$,
\begin{equation*}
\Delta_{rMSE}(i,w_1,w_2) = rMSE(i,w_1) - rMSE(i,w_2)
\end{equation*}
We also report between-group disparities in readmission cost (normalized by number of discharges per group) and in baseline-adjusted readmission cost, $c_B$, which subtracts each patient's readmission risk at the maximum LOS. This adjustment accounts for group differences in baseline risk. %

\begin{figure*}[ht]
\centering
\includegraphics[width=1.0\textwidth]{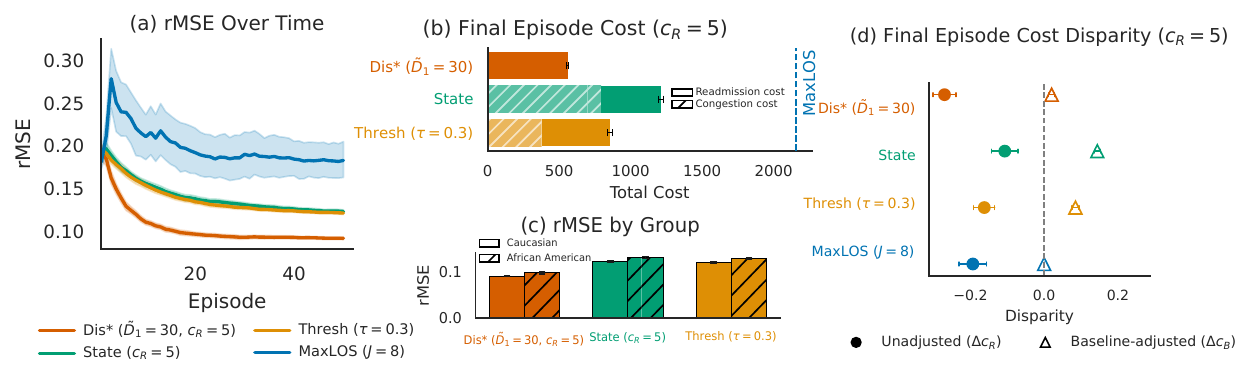}
\caption{Performance of the discharge up to policy compared to best-performing baselines in a simulation calibrated with real hospital readmission data. (a) rMSE of all model iterates with bootstrapped 95\% confidence intervals. (b) Total cost of each policy broken down into readmission and congestion cost, for each cost regime. SE bars are for the total cost. (c) $rMSE$ by group (Caucasian and African American) at the final episode, for the best-performing parameter of each policy class. (d) Readmission disparities and baseline-readjusted disparities (Caucasian - African American) at the final episode in each cost regime. }\label{fig:realistic}
\end{figure*}
\paragraph{Uncertainty Quantification.} To address variance from stochasticity in arrivals and readmissions, we perform 25-50 Monte Carlo simulations for each experiment. We test for significance of improvements with a one-sided $t$-test (testing for improvement in absolute value for $rMSE$ and cost disparities), with Bonferroni-corrected rejection level $\alpha=0.005$. Sample means and SEs for all metrics are in Appendix \ref{app:simulation:results}.

\subsection{Simple Setting: Exploration and Learning}

In our simple setting, the discharge-up-to policy consistently achieves the lowest rMSE (Figure \ref{fig:rmse}b). Across different risk profiles, model initializations, and discharge levels, stochasticity in the arrivals process drives sufficient exploration to support learning. By contrast, baseline policies fail to learn when they concentrate observations at a narrow set of LOS values. Simple threshold policies only explore when the decision threshold $\tau$ lies in a region where small model updates significantly affect which LOS gets discharged (Figure \ref{fig:rmse}c). For state-aware threshold policies, increasing readmission cost $c_R$ leads to censoring of shorter-stay outcomes, eventually approximating the maximum LOS policy (Figure \ref{fig:rmse}d).

\begin{table*}[t]
\centering
\small
\setlength{\tabcolsep}{4pt}
\begin{tabular}{lcccc}
\hline
Method & rMSE & $|\Delta rMSE|$ & Cost & $|\Delta c_B|$ \\
\hline
Dis* ($\tilde{D}_1=30, c_R=5$) & 0.095 & 0.0113 & 558.9974 & 0.0538 \\
\hline
Thresh & 0.1212 ($\tau=0.3$) & 0.0110 ($\tau=0.2$) & 854.8178 & 0.0862 ($\tau=0.1$) \\
Improvement over Thresh
& 24.4\% ($p<.001$)
& -2.49\% ($p=.59$)
& 34.4\% ($p<.005$)
& 40.3\% ($p<.005$) \\
\hline
State ($c_R=5$) & 0.123 & 0.0114 & 1212.6527 & 0.1439 \\
Improvement over State
& 25.5\% ($p<.005$)
& 0.8\% ($p=.48$)
& 37.7\% ($p<.005$)
& 62.7\% ($p<.005$) \\
\hline
\end{tabular}
\caption{Comparison of performance of discharge-up-to policy at the final episode $c_R=5$ against best-performing baselines.}
\label{tab:discharge_results}
\end{table*}

\subsection{Realistic Performance}
In our realistic setting, the discharge-up-to policy achieves lowest $rMSE$ by the second retraining episode. In the final episode, the best-performing discharge-up-to policy reduces error over the best-performing state aware policy and threshold policy by $32\%$ (Figure \ref{fig:realistic}a). In a congestion-focused cost environment ($c_R=5$), the discharge-up-to policy also reduces cost over the best baselines by at least $35\%$ (Figure \ref{fig:realistic}b). By contrast, when readmission costs are high ($c_R=125$) and optimal behavior approaches the maximum LOS policy, the discharge-up-to policy matches the baselines.

Improved estimation under the discharge-up-to policy is mixed at improving disparities in learning; while overall $rMSE$ improves for both groups (Figure \ref{fig:realistic}c), there is no improvement in absolute $rMSE$ disparities (Table \ref{tab:discharge_results}). Still, overall improvements in $rMSE$ contribute to ameliorating some forms of decision bias. By the final period, the discharge-up-to policy reduces baseline-adjusted readmission cost disparities by at least 57\% (Figure \ref{fig:realistic}d), but African American patients have similar unadjusted readmission costs across policies, reflecting higher baseline risk. Interestingly, the direction of adjusted disparity reduced is for Caucasian patients, perhaps reflecting differing benefits of longer stays; African-American patients may have lower marginal benefits due to structural inequalities e.g. in access to ongoing care. An equity-focused intervention should couple marginal risk discharging with interventions reducing structural factors. In the interim, improving adjusted disparities may benefit African-American patients as longer-than-necessary stays impose financial cost and risks like hospital-acquired complications.

\section{Discussion}

Simulations suggest that the asymptotic learning advantage of discharge-up-to policies (Theorem~\ref{thm:main}, Corollary~\ref{cor:wellspec}) matters in finite samples and in more realistic hospital settings, reducing prediction error, cost and disparities in prediction error, and baseline-adjusted readmission cost. At the same time, improved predictive fairness does not automatically imply equality in unadjusted realized readmission costs, as readmission disparities also reflect baseline health differences, resulting from social determinants of health and post-discharge resources. Thus, capacity-aware learning is one component of fair discharge policy, and should work alongside broader interventions to reduce readmission disparities.

Several limitations point to useful future work.  Daily optimization of the discharge up to level, as implemented in \citet{shi2021timing}, could improve both cost and estimation by diversifying discharge decisions. Our theoretical analysis relies on a stylized queueing model and either correct model specification or known reweighting adjustments under misspecification. Real deployments would require estimating discharge probabilities for IP-weighting under model misspecification, accounting for unmeasured covariates, clinician discretion and post-discharge interventions, as well as validating that longer stays are clinically appropriate and cost-effective for the patients who receive them. More broadly, our results suggest that fairness interventions in healthcare should be designed at the level of the coupled learning-and-decision system. Operational variability that already exists in hospitals can sometimes be used to ameliorate censoring bias, but it must be governed carefully so that learning ultimately improves care for the most vulnerable patients.

\bibliography{mybib}

\clearpage
\appendix
\onecolumn

\section*{Appendix}

\section{Definitions and Model}\label{app:model}

\subsection{Notation}\label{app:notation}

\begin{table}[H]
\centering
\caption{Notation table.}\label{tab:notation}
\begin{tabular}{p{0.18\textwidth} p{0.75\textwidth}}
\toprule
Notation & Description \\
\midrule
$\|\cdot\|_1$, $\|\cdot\|$ & Entrywise $L_1$ norm \\
$A_{k,w}$ & Number of arrivals on day $k$ for patients of type $w$ \\
$A_k$ & Number of arrivals of all types, $\sum_{w\in\mathcal{W}} A_{k,w}$ \\
$c_B(D_k)$ & Expected readmission cost of discharges adjusted for baseline risk (risk if held until maximum LOS), $c_R\sum_{w=1}^M \sum_{\ell=1}^J D_{k,w}^\ell \cdot (r_w(\ell)-r_w(J))$ \\
$c_B(D_k,w)$ & Expected readmission cost of discharges from group $w$ adjusted for baseline risk $c_R \sum_{\ell=1}^{J} D_{k,w}^{\ell} r_w(\ell)$ \\
$\Delta_{c_B}(i,w_1, w_2)$ & Disparity in adjusted expected readmission cost between $w_1$ and $w_2$ for round $i$. See Appendix \ref{app:simulations:methods} for definition.\\
$c_h(S_k,D_k)$ & Congestion cost of a patient census $S_k$. May be measured pre-action or post-action after discharging $\|D_k\|_1$ patients. Example forms are quadratic $h \cdot S_k^2$ or piecewise linear overage $h\cdot \max(0,S_k-N)$\\
$c_R$ & Scalar cost of a readmission to the hospital \\
$c_R(D_k)$ & Expected readmission cost from discharge matrix $D_k$, $c_R\sum_{w=1}^{M}\sum_{\ell=1}^{J} D_{k,w}^{\ell} r_w(\ell)$ \\
$c_R(D_k,w)$ & Expected readmission cost of discharges from group $w$, $c_R \sum_{\ell=1}^{J} D_{k,w}^{\ell} r_w(\ell)$ \\
$\Delta_{c_R}(i,w_1, w_2)$ & Disparity in expected readmission cost between $w_1$ and $w_2$ for round $i$. See Appendix \ref{app:simulations:methods} for definition.\\
$\mathcal{D}$ & Dataset of patient features, LOS, and readmission outcomes -- for each patient $i$, $(z^e_i,\ell_i,Y_i)$  where $z^e$ collects patient covariates such as group labels $w_i$\\
$\mathcal{D}_t$ & Dataset of patient features, LOS and readmission outcomes collected in episode $t$ \\
$\mathcal{D}_w$ & Dataset of patient features, LOS and readmission outcomes restricted to patients in group $w$ $\{(z^e_i, \ell_i, Y_i) : w_i=w\} $\\
$D_{k,w}^{\ell}$ & Number of discharged type-$w$ patients with LOS $\ell$ on day $k$ \\
$D_{k,w}$ & Vector of patient-type $w$ discharges on day $k$, $(D_{k,w}^{1},\ldots,D_{k,w}^{J})$ \\
$D_k$ & Matrix in $\mathbb{N}^{M\times J}$ of patient discharge vectors $D_{k,w}$ for each $w\in\mathcal{W}$ \\
$h$ & Scalar parameter for the holding cost $c_h$ \\
$J$ & Maximum LOS \\
$\lambda$ & Patient arrival rate  \\
$\lambda_w$ & Arrival rate for group $w$, $\lambda\pi_w$ \\
$M$ & Number of patient groups \\
$N$ & Hospital capacity in overage congestion cost, $h\cdot \max(0, S_k-N)$\\
$\pi_w$ & Proportion of arriving patients that are type $w$, $\sum_{w\in\mathcal{W}}\pi_w=1$ \\
$r_w$ & True readmission risk for type-$w$ patients\\
$r$ & True readmission risk (no patient type)\\
$rMSE(i)$ & $rMSE$ on retraining round $i$. See Appendix \ref{app:simulations:methods} for definition. \\
$rMSE(i,w)$ & $rMSE$ for group $w$ on retraining round $i$. See Appendix \ref{app:simulations:methods} for definition. \\
$\Delta_{rMSE}(i,w_1, w_2)$ & Disparity in root mean squared error between $w_1$ and $w_2$ for round $i$. See Appendix \ref{app:simulations:methods} for definition.\\
$\hat{r}_{w,t}$ & Estimated readmission risk for type-$w$ patients at retraining round $t$ \\
$S_{k,w}$ & Total patient census for type $w$ on day $k$, $\sum_{\ell=1}^{J} X_{k,w}^{\ell}$ \\
$S_k$ & Total patient census on day $k$, $\sum_{w=1}^{M} S_{k,w}$ \\
$T$ & Total number of days per episode \\
$\mathcal{W}$ & Set of patient types \\
$X_{k,w}^{\ell}$ & Number of type-$w$ patients with LOS $\ell$ in the hospital on day $k$ \\
$X_{k,w}$ & State vector of type-$w$ patients on day $k$, $(X_{k,w}^{1},\ldots,X_{k,w}^{J})$ \\
$X_k$ & Matrix in $\mathbb{N}^{M\times J}$ of patient state vectors $X_{k,w}$ for each $w\in\mathcal{W}$ \\
$z$  & Patient covariates $(\log(\ell), z^e)$  \\
$z^e$ & Policy-exogenous patient covariates \\
\bottomrule
\end{tabular}
\end{table}

\subsection{Baseline Threshold Policy Definitions}\label{app:threshold-policies}

We compare against two threshold-based baselines that reflect the status quo use of risk scores in discharge decisions \citep[see, e.g.,][]{shi2021timing}. The \emph{simple threshold policy} $\Pi_\tau$ discharges all patients whose estimated risk $\hat{r}_w(\ell)$ is below a fixed threshold $\tau$. The \emph{myopic state-aware threshold policy} $\Pi_s$ generalizes this by adapting the threshold to the current census: it greedily discharges patients in increasing order of estimated risk (i.e., starting with lowest risk patients) until the marginal readmission cost exceeds the marginal congestion cost at the current occupancy level.

\paragraph{Simple threshold policy.}
The simple threshold policy $\Pi_\tau$ discharges all patients whose estimated readmission risk falls below a fixed threshold $\tau$:
\begin{equation}
D_{k,w}^\ell = \begin{cases}
X_{k,w}^\ell & \hat{r}_w(\ell) < \tau \\
0 & \text{otherwise,}
\end{cases}
\end{equation}
for each type $w$ and length of stay $\ell < J$.

\paragraph{Myopic state-aware threshold policy.}
The myopic state-aware threshold policy $\Pi_s$ adapts the discharge threshold to the current patient census (after discharging all patients at the max LOS). When the census is $S$ and the congestion cost takes quadratic form $h \cdot S^2$, the upper risk threshold after which the marginal cost of discharging a patient exceeds its marginal contribution to the congestion cost is:
\begin{equation}
\tau(S) = \begin{cases}
\frac{h}{c_R}(S + 1) & S > 0 \\
0 & \text{otherwise.}
\end{cases}
\end{equation}

When the holding cost takes the piecewise linear form $\max(0,S-N)$, the threshold is:

\begin{equation}
\tau(S) = \begin{cases}
\frac{h}{c_R} & S - N > 0 \\
0 & \text{otherwise.}
\end{cases}
\end{equation}

Policy $\Pi_s$ discharges patients across all types in increasing order of estimated risk $\hat{r}_w(\ell)$, applying the threshold $\tau(S)$ and decrementing $S$ after each discharge. Formally, let
\[
T_k^{w,\ell} = \sum\{ X_{k,w'}^j : (w',j) \text{ where } \hat{r}_{w'}(j) < \hat{r}_w(\ell) \} \cup \{X_{k,w}^J : w \in W\}
\]
be the total number of patients (across all types) with lower estimated risk than a type-$w$ patient at length of stay $\ell$ as well as patients at the maximum LOS, who are discharged automatically. Then:
\begin{equation}
D_{k,w}^\ell = \max_d\ \hat{r}_w(\ell) < \tau(S_k - T_k^{w,\ell} - d') \quad \forall\, d' \in [0, d-1].
\end{equation}

From the perspective of each individual discharge decision, $\Pi_s$ is cost-optimal, finding the lowest cost patient to discharge among those whose expected readmission cost is below their marginal contribution to the congestion cost. For the piecewise linear congestion cost (if assessed post-action as $\max(S_k-||D_k||-N,0)$), the policy is also cost-optimal per day; the $N$ ``free'' congestion cost slots are reserved for the highest-risk patients, and then  all remaining patients with lower marginal readmission than congestion cost are discharged. However, the policy does not account for the downstream effects of discharge decisions on future congestion. The discharge-up-to policy (\S\ref{sec:theory:discharge_policies}) addresses this limitation.

\subsection{Strong Dominance and Priority Orders}\label{app:dominance}

\begin{definition}[Strong dominance]\label{def:strong_dominance}
A priority ordering $\succ$ satisfies \emph{strong dominance} if, for all $(w, \ell) \succ (w', \ell')$, extending the stay of a type-$w$ patient from $\ell$ to $\ell + j$ days yields a weakly smaller marginal risk reduction than extending a type-$w'$ patient from $\ell'$ to $\ell' + j$ days, for every possible extension $j$:
\begin{equation}
(w, \ell) \succ (w', \ell') \implies \hat{r}_w(\ell) - \hat{r}_w(\ell + j) \leq \hat{r}_{w'}(\ell') - \hat{r}_{w'}(\ell' + j) \quad \forall\, j \in [1, \min(J-\ell, J-\ell')].
\end{equation}
\end{definition}

\begin{fact}[Existence of strong dominance rankings]\label{fact:strong_dom_M1}
For $M=1$, diminishing marginal returns of hospitalization, that is, $\hat{r}(\ell) - \hat{r}(\ell+1)$ is non-increasing in $\ell$ over $[1,J]$, is sufficient for the existence of a ranking satisfying strong dominance.
\end{fact}
\begin{proof}
The ordering $\succ$ that ranks patients in decreasing order of length of stay satisfies strong dominance by definition.
\end{proof}

\begin{fact}[Checking for Strong Dominance for $M\ge 2$]\label{fact:strong_dom_M2}
More generally, for $M \geq 2$, a strong dominance ranking over pairs $(w, \ell)$ exists for risk functions $\{\hat{r}_w\}_{w \in W}$ if for every pair $(w, \ell)$ and $(w', \ell')$, the strong dominance condition holds in at least one direction: either
\begin{equation}
\hat{r}_w(\ell) - \hat{r}_w(\ell+j) \leq \hat{r}_{w'}(\ell') - \hat{r}_{w'}(\ell'+j) \quad \forall\, j \in [1, \min(J-\ell, J-\ell')],
\end{equation}
or the reverse inequality holds for all such $j$.\footnote{Verifying this requires $O(M^2 J^3)$ checks in total.} This ensures $\succ$ is complete, yielding a total order over all $(w, \ell)$ pairs.
\end{fact}
\begin{proof}This follows from Definition~\ref{def:strong_dominance}.\end{proof}

\begin{figure}[ht]
    \centering
    \includegraphics[scale=0.3]{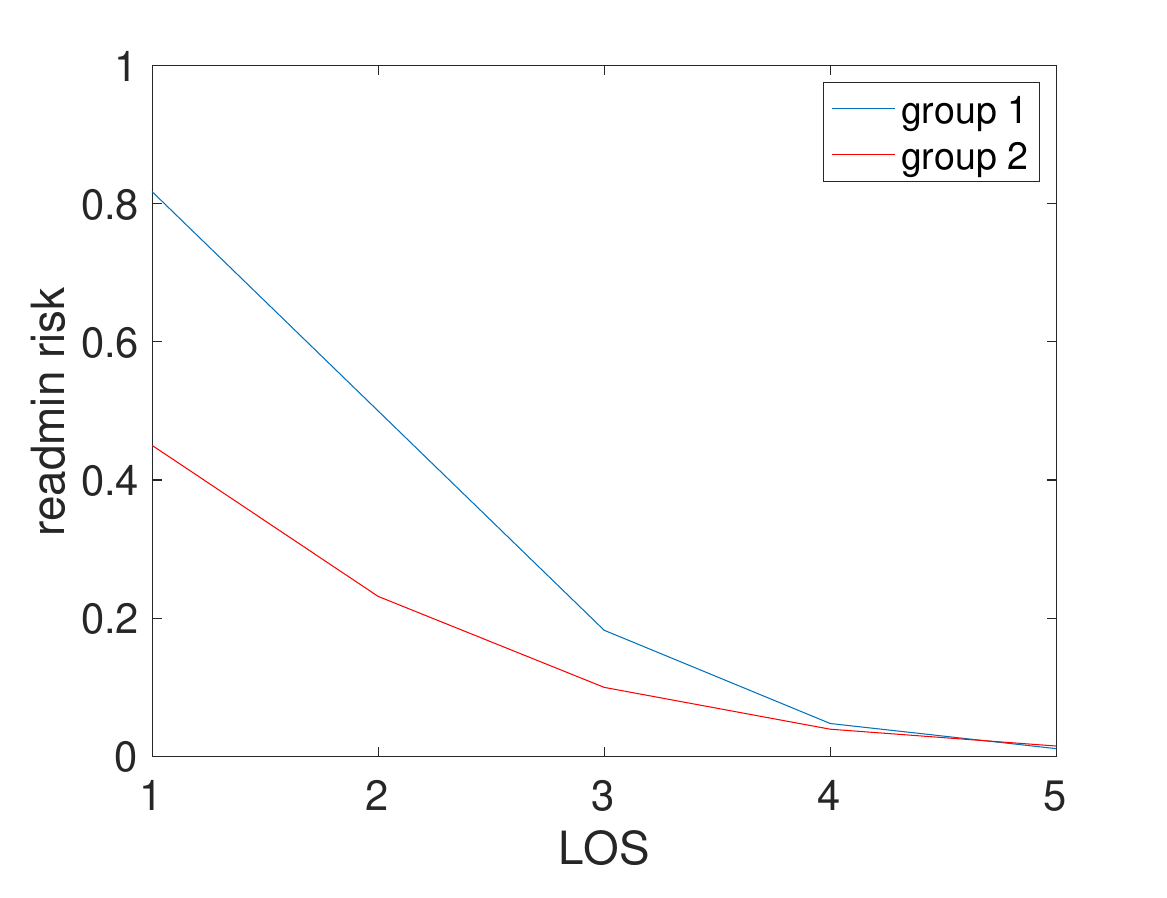}
    \caption{Readmission risk as a function of LOS for two groups. For group 1, the logit $\ln(r(\ell)/(1-r(\ell)))$ follows $-1.6\ell + 3$; for group 2, it follows $-\ell + 0.8$. Group 1 has higher baseline risk and larger marginal risk reductions from continued hospitalization; group 2 has lower baseline risk and smaller marginal reductions.}
    \label{fig:readm_example}
\end{figure}

\paragraph{Illustrative example.}
We illustrate the strong dominance priority ranking with two patient types, $W = \{1, 2\}$. For each type $w$, readmission risk follows the logistic form~\eqref{eq:logit-risk} with type-specific parameters. Figure~\ref{fig:readm_example} shows type 1 as higher-risk with a steeper $\hat{r}_1(\ell)$ curve and type 2 as lower-risk with a shallower $\hat{r}_2(\ell)$ curve.

Starting from LOS four, the strong dominance ranking (most to least preferred for discharge) is:
\begin{equation}
(2,4) \succ (1,4) \succ (2,3) \succ (1,3) \succ (2,2) \succ (2,1) \succ (1,2) \succ (1,1).
\end{equation}
Under this ranking, type-2 patients are generally favored for earlier discharge because their marginal benefit from additional hospitalization is smaller, resulting in shorter average LOS for that type.

\subsection{Priority Ordering by Weak Dominance}\label{app:weak_priority_order}

The optimality result discussed in Section~\ref{sec:theory:discharge_policies} uses strong dominance to justify the discharge-up-to form. In practice, we may use a weaker priority rule based on estimated average marginal risk reductions.

For each type $w$ and length of stay $\ell < J$, define the score
\begin{equation}
\Omega_w(\ell) = \frac{1}{J - \ell} \sum_{j=1}^{J-\ell} \bigl(\hat{r}_w(\ell) - \hat{r}_w(\ell + j)\bigr),
\end{equation}
which measures the average marginal risk reduction over all possible further lengths of stay for type $w$. A ranking satisfies \emph{weak dominance} when pairs are ordered by increasing score.

\begin{definition}[Weak dominance]\label{def:weak_dominance}
A priority ordering $\succ$ satisfies \emph{weak dominance} if pairs are ranked by increasing average marginal risk reduction:
\begin{equation}
(w, \ell) \succ (w', \ell') \implies \Omega_w(\ell) \leq \Omega_{w'}(\ell').
\end{equation}
\end{definition}
 
The intuition is that patients with low $\Omega_w(\ell)$ have little to gain from further hospitalization on average and should be prioritized for discharge. Strong dominance implies weak dominance because it requires the relevant marginal risk inequality to hold for every possible extension length $j$, rather than only on average. 

\section{Proofs}

\subsection{Supplemental Analysis of Discharge-Up-To Policy}\label{app:discharge-up-to}

The most important finding from~\citet{shi2021timing} is that the daily discharge decision can be decomposed into two steps: first, decide on a ranking order to determine which patient to be prioritized for discharge; second, based on the ranking, determine the univariate decision variable $\norm{D_k}_1$ on the total number of patients to discharge. 

To gain structural insights, we consider a special class of discharge policy, assuming that we decouple the two steps: We decide on the total number of discharges $\norm{D_k}_1$ each day without considering the actual patient status (e.g., how long the patients have stayed). In other words, we choose patients at random to discharge until we achieve the number $\norm{D_k}_1$. For traceability, we assume that there exists a distribution $\{q_j\}$'s, independent of the previous actions, that characterizes the LOS distribution of the patients in the system in each epoch. As a result, the discharge cost can be simplified as 
$$ c_R \cdot \sum_{j} q_j r(j) \norm{D_k}_1 \triangleq R \norm{D_k}_1 $$
by setting $R =c_R \cdot \sum_{j} q_j r(j)$. 

\begin{proposition}\label{prop:optimal}
Under the given class of discharge policies and a quadratic form of the value function, the simple \emph{discharge-up-to} policy is optimal. 
\end{proposition}

\begin{proof}
    To set up the analysis, we use backward induction. 
For the last period $k=T$, the value function automatically satisfy the quadratic form as the cost equals $h(S_T - \norm{D_k}_1)^2 + R \tilde{D}_T$. 
Now suppose the value function for period $k+1$ follows the form of $V(S_k) = \alpha_{k+1} S_{k+1}^2 + \beta_{k+1} S_{k+1} + \gamma_{k+1}$. Then, we have the Bellman equation as 
\begin{equation}
V(S_k) = \min_{\norm{D_k}_1} \quad 
h(S_k - \norm{D_k}_1)^2 + R \norm{D_k}_1 
+ \mathbb{E}[V(S_k - \norm{D_k}_1 + A_k)] . 
\end{equation}
Plugging in the quadratic form of the value function for $k+1$, we get that 
\begin{eqnarray}
f(\norm{D_k}_1) &=& h(S_k - \norm{D_k}_1)^2 + R \norm{D_k}_1 
+ \alpha_{k+1} (S_k - \norm{D_k}_1)^2 + 2\alpha_{k+1}\mathbb{E}[A_k] (S_k - \norm{D_k}_1) 
\\
&&{} 
+ \alpha_{k+1} \mathbb{E}[A_k^2] 
+ \beta_{k+1} (S_k - \norm{D_k}_1) + \beta_{k+1} \mathbb{E}[A_k] + \gamma_{k+1},  
\end{eqnarray}
where $\mathbb{E}[A_k^2]$ and $\mathbb{E}[A_k]$ do not depend on $\norm{D_k}_1$. As a result, taking the derivative of $f(\norm{D_k}_1)$, we get that 
\begin{eqnarray*}
f^\prime (\norm{D_k}_1) &=& -2 (h + \alpha_{k+1}) (S_k - \norm{D_k}_1) + (R - 2\alpha_{k+1}\mathbb{E}[A_k] - \beta_{k+1}).      
\end{eqnarray*}
Setting the derivative to be zero, we have the optimal discharge decision
\begin{equation}
S_k - \norm{D_k}_1^* =   \frac{R - 2\alpha_{k+1}\mathbb{E}[A_k] - \beta_k}{2 (h + \alpha_{k+1})} 
\label{eq:optimal-discharge}
\end{equation}
when $R - 2\alpha_{k+1}\mathbb{E}[A_k] - \beta_{k+1} \in (0, S_k)$; otherwise $\tilde{D}^*_k$ reaches the boundary (discharge all or nothing). Equation~\eqref{eq:optimal-discharge} says that the
optimal discharge quantity should bring the occupancy level to a constant level that is determined by $R - 2\alpha_{k+1}\mathbb{E}[A_k] - \beta_k$ and $h + \alpha_{k+1}$. In other words, the optimal policy follows the ``discharge-up-to'' structure. \hfill 
\end{proof}

\medskip 

The interpretation of this policy structure is also intuitive:
\begin{itemize}
    \item If the readmission penalty $R$ is large, the constant level is high, meaning that fewer patients should be discharged (to reduce the readmission) and the occupancy is maintained at a higher level;
    \item If the holding cost $h$ is large, the constant level is low, meaning that more patients should be discharged to reduce the occupancy level and avoid unit congestion. 
\end{itemize}

The discharge-up-to policy is shown to be optimal under strong conditions, which ignore the current patient status. In other words, this policy can be seen as a steady-state control (assuming the patients' LOS distribution has reached a steady-state distribution), while it is not necessarily the optimal policy under a fully dynamic control setting. 

\subsection{Distribution support failure under deterministic arrivals}\label{app:bias-in-bias-out}

\paragraph{Setup}
Consider two patient types \(a,b\) with constant arrivals $A_{k,a}=n_a,
A_{k,b}=n_b$, for all $k$,
where \(n_a,n_b\ge 1\). Fix \(1\le L_b<L_a<J\), a retraining
round \(t\), and the priority ordering \(\succ_t\) induced by the
currently deployed predictor.

For \(d_a\in\{0,\ldots,n_a\}\), set $\widetilde D
=
n_a(L_a-1)+n_b(L_b-1)+d_a.$

Suppose the priority ordering satisfies
\begin{align}
(b,L_b)&\succ_t(a,L_a), \label{eq:det-ranking-b}\\
(a,L_a)&\succ_t(a,\ell)
&&\text{for all }1\le \ell<L_a, \label{eq:det-ranking-a}\\
(a,L_a)&\succ_t(b,\ell)
&&\text{for all }1\le \ell<L_b, \label{eq:det-ranking-lower}
\end{align}
such that group $b$ is discharged first.
If \(d_a>0\) and \(L_a+1<J\), also suppose $(a,L_a+1)\succ_t(a,L_a)$.
If \(L_a+1=J\), the patients at \(L_a+1\) are automatically
discharged.

Let \(x^{d_a}\) denote the state with \(n_a\) type-\(a\) patients at
each LOS \(1,\ldots,L_a\), an additional \(d_a\) type-\(a\) patients
at LOS \(L_a+1\), and \(n_b\) type-\(b\) patients at each LOS
\(1,\ldots,L_b\), with all other cells empty.

\begin{proposition}[Distribution support failure under deterministic
arrivals]
\label{prop:deterministic-support}
Under the discharge-up-to policy \(\Pi_{\widetilde D}\) with priority
ordering \(\succ_t\), if \(X_k=x^{d_a}\), then
$X_{k+1}=x^{d_a}$.
On each such day,
$D^{L_a+1}_{k,a}=d_a,
D^{L_b}_{k,b}=n_b,
D^{L_a}_{k,a}=n_a-d_a$,
and no other patients are discharged. Consequently, repeated
application of this operational policy produces no type-\(b\)
observations beyond \(L_b\); in the notation of Section~5,
\[
d^{\Pi_{\widetilde D}}(\ell,b)=0
\qquad
\text{for every }\ell>L_b.
\]
\end{proposition}

\begin{proof}
At \(x^{d_a}\), the total census is $S_k=n_aL_a+n_bL_b+d_a$.
Thus the policy discharges $S_k-\widetilde D=n_a+n_b$
patients. The priority ordering removes the \(d_a\) type-\(a\)
patients at \(L_a+1\) and the \(n_b\) type-\(b\) patients at \(L_b\)
before reaching \((a,L_a)\). The remaining \(n_a-d_a\) discharges
are then taken from \((a,L_a)\), so the target is reached before any
lower-LOS cell is considered.

After the state update and deterministic arrivals, the resulting state is again
\(x^{d_a}\). 
The support claim follows because no type-\(b\) patient
ever reaches an LOS greater than \(L_b\).
\end{proof}

\subsection{Proof of Theorem~\ref{thm:main}}\label{app:non-censorship-proof}

\begin{proof}
Fix an arbitrary patient $i$ that arrives on day $k$, so that patient $i$ has $\mathrm{LOS}=1$ at the
start of day $k+1$. Let $L$ denote the final LOS of patient $i$, that is, the patient $i$ is discharged on day $k+L$.

We show that for each $j\in\{1,\dots,J\}$ there exists an event with strictly positive probability on which
$L=j$. The proof proceeds by induction on $j$ via positive conditional probabilities.

\paragraph{Base case ($j=1$)} Since patients cannot be discharged on the day they arrive, we must have $\Pr(L\ge 1)>0$.

\paragraph{Induction step.} Fix $\ell\in\{1,\dots,J-1\}$ and suppose $\Pr(L\ge \ell)>0$. We show that both
$\Pr(L=\ell)>0$ and $\Pr(L\ge \ell+1)>0$.

Condition on the event $\{L\ge \ell\}$, i.e., patient $i$ survives through the end of day $k+\ell-1$ and has
$\mathrm{LOS}=\ell$ at the start of day $k+\ell$.

\emph{(a) Positive probability of discharge at $\ell$.}
Consider the event that the system on day $k+\ell$ is sufficiently congested so that the discharge-up-to policy must discharge
at least one patient who has the same discharge priority as patient $i$. We analyze two cases separately.

First suppose $\ell =1$ and patient $i$ is type $w^*$.
On the event $E = \{A_{k,w^*} = c+1\}$, the system has $c+1$ patients of type $w^*$ with $\textrm{LOS} = 1$ on the start of day $k+1$ (prior to discharge decisions); hence at least one patient of type $w^*$ with $\textrm{LOS} = 1$ must be discharged in order to bring the occupancy down to $c$. Under the random tie-breaking assumption, patient~$i$ is selected to be discharged with probability $1/(c+1)$ on the event $E$. Therefore,
\begin{equation*}
    \Pr(L=1) \ge \frac{1}{c+1}\cdot\Pr(A_{k, w^*} = c+1 ) > 0,
\end{equation*}
where $\Pr(A_{k,w^*} = c+1 ) > 0$ holds by the positive support assumption on the arrival process.

Now suppose either $\ell \geq 2$ or patient $i$ is not type $w^*$. On the event $E = \{A_{k+\ell - 1,w^*} = c\}$, the system has at least $c+1$ patients, including $c$ type $m^*$ patients with $\textrm{LOS} = 1$ and patient $i$, on the start of day $k + \ell$ prior to discharge decisions. Hence, at least one patient must be discharged in order to bring the occupancy down to $c$. Under the ``weak preferences for new arrivals'' assumption, patient $i$ must be discharged with probability at least $1/(c+1)$ on event $E$. Therefore,
\begin{equation*}
    \Pr(L=\ell \mid L\ge \ell) \ge \frac{1}{c+1}\cdot \Pr(A_{k+\ell-1,w^*}=c) >  0,
\end{equation*}
so we have $\Pr(L=\ell)>0$.

\emph{(b) Positive probability of survival beyond $\ell$.} By the discharge-up-to policy on day $k+\ell-1$, we must have $S_{k+\ell-1} - \tilde{D}_{k+\ell-1}\le c$. Consider the event $E = \{A_{k+\ell-1}=0\}$. On this event, the system's occupancy on day $k+\ell$ is computed as $S_{k+\ell} =  S_{k+\ell-1} - \tilde{D}_{k+\ell-1} +A_{k+\ell-1} \le c$. According to the discharge-up-to policy, there will be no discharges on day $k+\ell$. This implies that patient $i$ survives to day $k+\ell+1$ and that $L \ge \ell+1$ on the event $E$. We obtain
\begin{equation*}
    \Pr(L\ge \ell+1 \mid L\ge \ell) \ge \Pr(A_{k+\ell-1}=0)> 0,
\end{equation*}
so we have $\Pr(L\ge\ell+1)>0$.

By repeated application of the induction step, we have $\Pr(L=j) > 0$ for every $j \in \{2, \cdots, J-1\}$. Moreover, we have $\Pr(L\ge J) > 0$. Since any patient that reaches LOS$=J$ is discharged by the end of that day (maximum length of stay), we have $\Pr(L = J) = \Pr(L\ge J) > 0$.
    This completes the proof.
\end{proof}

\section{Simulations}\label{app:simulations}

\subsection{Methods} \label{app:simulations:methods}

\paragraph{Models}  We construct 10 true risk models in the fully simulated setting  by varying risk on the first day $r(1)$ and the maximum LOS $r(J)$, and solving for intercept $b_0$ and LOS coefficient $b_1$ assuming logistic functional form (Equation \ref{ex:logistic}). We sweep over $r(1)\in[0.6, 0.7, 0.8, 0.9]$ and $r(J)\in [0.3, 0.4,\ldots, r(1)-0.3]$. We construct five starting risk models $\hat{r}_1$ for each true model $r$ by taking all combinations of $\hat{r}(1)=r(1)\pm 0.1$ and $\hat{r}(J)=r(J)\pm 0.1$, as well as one setting where estimated risk is equal to true risk $\hat{r}(1)=r(1), \hat{r}(J)=r(J)$. This results in 50 total risk configurations, where each configuration is a paired true and starting risk model.

\begin{figure}[ht]
    \centering
\includegraphics[width=0.9\textwidth]{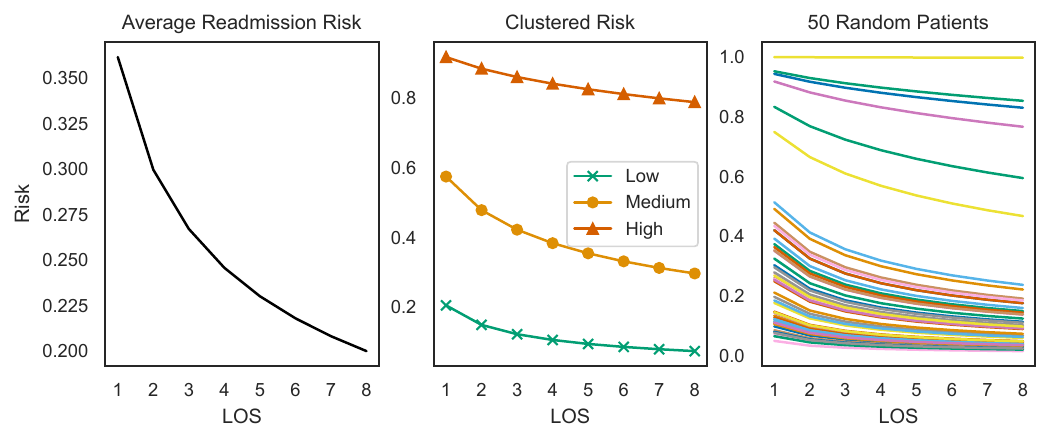}
    \caption{(a) Average risk over the empirical distribution of patients for the realistic model per LOS. (b) Average risk for three $k$-means clusters of patient risk trajectories. (c) Risk for 50 randomly sampled patients from our dataset.}
    \label{fig:realistic_risk}
\end{figure}

In the realistic setting, we use the EM algorithm detailed in \citet[Appendix A.2]{shi2021timing} to train a ground-truth readmission risk model. We train the model on $\frac{2}{3}$ of the  dataset; the model achieves $68.33\%$ AUC on the withheld test data. After the training process, we extract the stage 1 cure logistic model coefficients $\xi$ to construct the ground truth readmission model as $r(z)=1-\sigma(\xi z)=\sigma(-\xi z)$ with $z=(\log(\ell),z^e)$ for policy-exogenous patient features $z^e$. Figure \ref{fig:realistic_risk} visualizes the true risk model's predictions for patients in the dataset.

\begin{figure}[ht]
    \centering
    \includegraphics[width=0.5\linewidth]{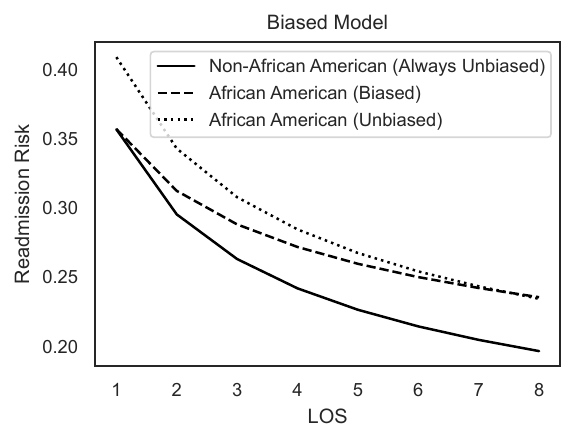}
    \caption{Average of biased model predictions vs. true risk predictions for the empirical distribution of  patient covariates conditional on race.}
    \label{fig:biased_risk}
\end{figure}

To construct the biased starting model $\hat{r}_1$, we multiply the LOS coefficient for African American patients' risk by $0.75$, which underestimates marginal risk,  and subtract $1.25$ from the coefficient encoding African-American race, which underestimates absolute risk. Figure \ref{fig:biased_risk} depicts the biased model's average predictions for African American patients in our dataset, as well as unbiased average risk predictions for African American and non-African American patients.

\subsubsection{Dataset} The dataset is pre-cleaned to exclude patients with planned readmissions, expired patients, patients under 18, obstetrics and gynecology patients. We additionally exclude readmitted patients (any arriving patient who was admitted within the previous 365 days). After these exclusions, the dataset includes $n=26,121$ patients ($23,609$ Caucasian, $2,242$ African-American, and $270$ other).

\paragraph{Policies} We choose policy parameters to fit each setting. In the fully simulated simple setting, we set our parameters as:
\begin{itemize}
\item \textbf{Discharge-up-to:} $\tilde{D}\in[10,20,\ldots,70]$
\item \textbf{Threshold:} $\tau \in [0.3, 0.4, 0.5]$
\item \textbf{State-aware:} $h=1$, $c_R\in [100,200,\ldots,1000]$
\end{itemize}

\noindent In the realistic setting, we set our parameters as:
\begin{itemize}
\item \textbf{Discharge-up-to:} The level is tuned with a grid search (described next), $\varepsilon=\frac{1}{10}$, $\mathcal{L}=[5,10,\ldots,55]$, $d\in [10,20,30]$, $h=1$, $c_R\in [5, 125]$, $\kappa=120$
\item \textbf{Threshold:} $\tau\in[0.1,0.2,0.3]$
\item \textbf{State-aware:} $h=1$, $c_R\in[5, 125]$

\end{itemize}

Our tuning procedure for the target discharge-up-to level is detailed in Algorithm \ref{alg:grid-optimization}. We start the  policy with an initial level $d$. At the end of each retraining episode $i$, with probability $\varepsilon$, we simulate the queue for $\kappa$ days at a set of candidate levels $\mathcal{L}$ and estimate cost according to $\hat{r}_i$, resetting the level to the one with minimal cost.  

\begin{algorithm}[tb]
\caption{Grid Search Discharge-Up-To Level Tuning}
\label{alg:grid-optimization}
\textbf{Input}: Candidate levels $\mathcal{L}$, initial level $d$, simulation length $\kappa$, exploration parameter $\varepsilon$, cost parameters $h,c_R$\\
\textbf{Output}: Tuned discharge level $\tilde{D}$
\begin{algorithmic}[1]
\STATE Set discharge level $\tilde{D} \gets d$
\STATE Set \texttt{exploredLevels} $\gets$ \textbf{False}
\STATE Initialize $\hat{r}^{(0)}$ to starting risk model

\FOR{episode $t = 1$ to $T$}
    \STATE Run the discharge-up-to policy with level $\tilde{D}$ for $k$ days
    \STATE Update risk model coefficients $\hat{r}_t$ using all observed readmissions up to and including episode $t$

    \IF{$\mathcal{U}(0,1) \leq \varepsilon$ }

        \FOR{each level $l \in \mathcal{L}$}
            \STATE Simulate discharge-up-to with level $l$ for $\kappa$ days using estimated model $\hat{r}_t$
            \STATE Estimate cost $C(l)$ using $\hat{r}_t$ to calculate the readmission cost
        \ENDFOR

        \STATE $\tilde{D} \gets \arg\min_{l \in \mathcal{L}} C(l)$
    \ENDIF
\ENDFOR

\STATE \textbf{return} $\tilde{D}$
\end{algorithmic}
\end{algorithm}

\paragraph{Cost Regimes}\label{app:parameters}
In the simple setting, we use the pre-discharge holding cost $h\cdot S_k^2$ to validate our theoretical results empirically. In the realistic setting, following \citet{shi2021timing}, we assess the pre-discharge overage cost $\max(S_k-N,0)$ with $N=20$ to probe empirical performance under a realistic cost function that is not theoretically tractable. 

Since the cost function and goals of the simulations differ, we test different $h:c_R$ ratios in our two settings, holding fixed $h=1$ and varying $c_R$. In the simple setting, we sweep over $c_R\in[100,200,\ldots,1000]$ to determine when state-aware threshold behavior approaches the maximum LOS policy. In the realistic setting, we test $c_R=[5,125]$, choosing these two settings to capture a congestion-focused vs. readmissions-focused cost regime respectively. Specifically, we analyze the tradeoff between average under the discharge-up-to policy run with the true risk model as we vary the level; from these curves, we pick $\tilde{D}=10$ as a realistic setting where congestion is low while readmissions are relatively high, while $\tilde{D}=30$ represents a case where congestion is relatively high and readmissions relatively low. We then select $c_R$ values such that these values of $\tilde{D}$ are evaluated as lowest-cost compared to the alternatives. 

\paragraph{Evaluation Metrics}  We evaluate the tool for learning, cost and disparities in both error and readmission cost.

We evaluate learning with \textbf{root mean squared error.} Because the stationary distribution of LOS differs by policy, we evaluate with respect to a reference distribution. In the simple setting, we evaluate the $rMSE$ over the uniform distribution of LOS:

\begin{equation}
rMSE(i) = \sqrt{\mathbb{E}_{\ell\sim \mathcal{U}(1,J)}[ (r(\ell)-\hat{r}_i(\ell)]}
\end{equation}

To evaluate rMSE in the realistic setting, we weigh other patient features according to the empirical covariate distribution from our dataset:
\begin{equation}
rMSE(i) = \sqrt{\mathbb{E}_{z^e\sim \mathcal{D},\ell\sim \mathcal{U}(1,J)}[ (r(\log(\ell),z^e) - \hat{r}(\log(\ell),z^e))^2 ]}
\end{equation}

We also examine \textbf{disparities in rMSE} across groups. In the real-data setting, groupwise $rMSE$ is average $rMSE$ according to the empirical distribution of covariates for patients in that group. With $\mathcal{D}^w=\{z^e_j\in \mathcal{D}: w_j=w \}$
\begin{equation}
rMSE(i,w) =  \sqrt{ \mathbb{E}_{z^e\sim \mathcal{D}^w,\ell\sim \mathcal{U}(1,J)} [(r(\log(\ell),Z^e) - \hat{r}_{i}(\log(\ell),Z^e) ^2}]
\end{equation}

The $rMSE$ disparity for two groups $w_1$ and $w_2$ is the difference in their groupwise $rMSE$:
\begin{equation}
\Delta_{rMSE}(i,w_1,w_2) = rMSE(i,w_1) - rMSE(i,w_2)
\end{equation}

We further evaluate policies based on \textbf{total cost}, which can be broken into \textbf{congestion cost} and \textbf{readmission cost}.  We use the cost functions defined in \S\ref{sec:decision_costs} and parameters described above (Cost Regimes).  We further break down the readmission cost into \textbf{the group readmission cost} for discharges into a particular group.

To measure disparities in patient cost, we consider the \textbf{readmission cost disparity}, defined as the absolute difference in the readmission cost per group, normalized by the number of discharged patients in each group (to account for different group sizes):

\begin{equation}
\Delta_{c_{R}}(i,w_1,w_2) =\abs[\Bigg]{\frac{\sum_{k=1}^{T_i} c_{R}(D_{k},w_1)}{\sum_k \norm{D_{k,w_1}}_1} - \frac{ \sum_{k=1}^{T_i} c_{R}(D_{k},w_2)}{\sum_k \norm{D_{k,w_2}}_1}}
\end{equation}

Readmission cost disparity can be viewed as a sufficiency notion \cite{barocas-hardt-narayanan}, measuring disparities in the true rate of readmission upon discharge across groups (scaled linearly by the constant cost to discharge $c_R$).

Finally, we consider a \textbf{baseline-adjusted readmission disparity} to account for baseline risk differences between group populations, by subtracting cost at the maximum possible LOS. We first define a baseline-adjusted readmission cost $c_B$, which in the simple setting takes the form:

\begin{equation}
c_B(D_k,w) = c_R \cdot \sum_{\ell=1}^J D_k^\ell ( r(\ell) -  r_w(J))\\
\end{equation}
$c_B$ is defined in the realistic setting by replacing groupwise predictions with personalized risk predictions $r(z)$. We then calculate the disparity $\Delta_{c_B}(i,w_1,w_2)$ for the baseline-adjusted readmission cost analogously to the readmission cost disparity. Baseline-adjusted readmission disparity can be viewed as checking for opportunity cost parity, or whether discharged patients across groups on average were discharged with equal opportunity to improve.

\paragraph{Uncertainty Quantification} We run 25 Monte Carlo simulations of each unique true/starting risk model configuration and policy for the simple setting, and 50 Monte Carlo simulations of each policy for the realistic data simulations. Our random seeds are 0-24 and 0-49 respectively.

For our realistic simulations in the congestion-focused setting $c_R=5$, we perform significance testing for any observed improvements/declines. For rMSE and rMSE disparity, we test for significance of improvement of the best-performing discharge policy vs. the best-performing baseline policies from each class (excluding the maximum LOS policy, which vastly under-performs other baselines and serves primarily as a conservative reference point). For total cost, readmission cost disparity, and baseline-adjusted readmission cost disparity, we test the best-performing baseline against the discharge-up-to policy with starting level $\tilde{D}=20$ (which is in the middle of our three starting levels and too high for the cost setting), to test the tuning algorithm's ability to find the cost-appropriate level. Together, this makes a total of 10 significance tests (5 metrics $\times$ 2 baseline classes) resulting in a Bonferroni-corrected rejection threshold of $\alpha=0.005$. 

\subsection{Results}\label{app:simulation:results}

For the simple setting, we report sample means and SEs of $rMSE$ and total cost at the final episode across configurations in Table \ref{tab:results_simple}. We report for a sample configuration $r(1)=\hat{r}(1)=0.7, r(J)=\hat{r}(J)=0.3$ in Table \ref{tab:results_simple_sample}. Sample means and SEs of all learning, cost, and disparity metrics are in Tables \ref{tab:real_rmse}-\ref{tab:real_cB}. Significance test results are in Table \ref{tab:significance}.

\begin{table}[ht]
\caption{Sample means and SEs for cost and $rMSE$ of each policy at the final episode across 50 true/initial configurations of risk.}\label{tab:results_simple}
\centering
\begin{tabular}{lrrrr}
\toprule
Policy Name & Mean Total Cost ($h=1, c_R=100$) & SEM Total Cost & Mean rMSE & SEM rMSE \\
\midrule
Dis ($\tilde{D}=10$) & $5.97 \times 10^{4}$ & 180.3612 & 0.0117 & 0.0002 \\
Dis ($\tilde{D}=20$) & $8.49 \times 10^{4}$ & 170.6799 & 0.0082 & 0.0002 \\
Dis ($\tilde{D}=30$) & $1.21 \times 10^{5}$ & 177.2092 & 0.0067 & 0.0001 \\
Dis ($\tilde{D}=40$) & $1.68 \times 10^{5}$ & 188.2904 & 0.0062 & 0.0001 \\
Dis ($\tilde{D}=50$) & $2.24 \times 10^{5}$ & 209.0363 & 0.0072 & 0.0001 \\
Dis ($\tilde{D}=60$) & $2.86 \times 10^{5}$ & 327.2783 & 0.0098 & 0.0002 \\
Dis ($\tilde{D}=70$) & $3.32 \times 10^{5}$ & 595.9752 & 0.019 & 0.0004 \\
MaxLOS ($J=8$) & $3.36 \times 10^{5}$ & 759.6766 & 0.1196 & 0.0014 \\
State ($c_R=100.0$) & $1.91 \times 10^{5}$ & 1205.7269 & 0.0088 & 0.0002 \\
State ($c_R=1000.0$) & NA & NA & 0.1196 & 0.0014 \\
State ($c_R=200.0$) & NA & NA & 0.0469 & 0.0016 \\
State ($c_R=300.0$) & NA & NA & 0.0918 & 0.0018 \\
State ($c_R=400.0$) & NA & NA & 0.1181 & 0.0015 \\
State ($c_R=500.0$) & NA & NA & 0.1177 & 0.0015 \\
State ($c_R=600.0$) & NA & NA & 0.1188 & 0.0015 \\
State ($c_R=700.0$) & NA & NA & 0.1196 & 0.0014 \\
State ($c_R=800.0$) & NA & NA & 0.1196 & 0.0014 \\
State ($c_R=900.0$) & NA & NA & 0.1196 & 0.0014 \\
Thresh ($\tau=0.3$) & $2.99 \times 10^{5}$ & 1682.7337 & 0.0829 & 0.0017 \\
Thresh ($\tau=0.4$) & $2.40 \times 10^{5}$ & 2433.1013 & 0.0489 & 0.0016 \\
Thresh ($\tau=0.5$) & $1.70 \times 10^{5}$ & 2528.2235 & 0.0117 & 0.0003 \\
\bottomrule
\end{tabular}
\end{table}

\begin{table}
\caption{Sample means and SEs for cost and $rMSE$ of each policy at the final episode for example risk configuration $r(1)=\hat{r}(1)=0.7$, $r(J)=\hat{r}(J)=0.3$.}\label{tab:results_simple_sample}
\centering
\begin{tabular}{lrrrr}
\toprule
Policy Name & Mean Cost ($h=1, c_R=100$) & SEM Total Cost & Mean rMSE & SE \\
\midrule
Dis ($\tilde{D}=10$) & $5.98 \times 10^{4}$ & 588.5656 & 0.0118 & 0.0015 \\
Dis ($\tilde{D}=20$) & $8.48 \times 10^{4}$ & 435.4185 & 0.0093 & 0.001 \\
Dis ($\tilde{D}=30$) & $1.21 \times 10^{5}$ & 652.0527 & 0.007 & 0.0006 \\
Dis ($\tilde{D}=40$) & $1.68 \times 10^{5}$ & 819.1144 & 0.0067 & 0.0009 \\
Dis ($\tilde{D}=50$) & $2.24 \times 10^{5}$ & 1006.7915 & 0.0076 & 0.0008 \\
Dis ($\tilde{D}=60$) & $2.86 \times 10^{5}$ & 2295.5058 & 0.0108 & 0.0014 \\
Dis ($\tilde{D}=70$) & $3.33 \times 10^{5}$ & 4209.9589 & 0.0219 & 0.0029 \\
MaxLOS ($J=8$) & $3.36 \times 10^{5}$ & 5473.2081 & 0.1174 & 0.0003 \\
State ($c_R=100.0$) & $1.91 \times 10^{5}$ & 1900.385 & 0.0087 & 0.0011 \\
State ($c_R=1000.0$) & NA & NA & 0.1174 & 0.0003 \\
State ($c_R=200.0$) & NA & NA & 0.0151 & 0.0023 \\
State ($c_R=300.0$) & NA & NA & 0.1173 & 0.0003 \\
State ($c_R=400.0$) & NA & NA & 0.1174 & 0.0003 \\
State ($c_R=500.0$) & NA & NA & 0.1174 & 0.0003 \\
State ($c_R=600.0$) & NA & NA & 0.1174 & 0.0003 \\
State ($c_R=700.0$) & NA & NA & 0.1174 & 0.0003 \\
State ($c_R=800.0$) & NA & NA & 0.1174 & 0.0003 \\
State ($c_R=900.0$) & NA & NA & 0.1174 & 0.0003 \\
Thresh ($\tau=0.3$) & $3.36 \times 10^{5}$ & 5473.2081 & 0.1085 & 0.0049 \\
Thresh ($\tau=0.4$) & $2.67 \times 10^{5}$ & 4220.4433 & 0.0247 & 0.0058 \\
Thresh ($\tau=0.5$) & $1.56 \times 10^{5}$ & 2389.4423 & 0.0105 & 0.0015 \\
\bottomrule
\end{tabular}
\end{table}

\begin{table}
\caption{Sample means and SEs for $rMSE$ and $rMSE$ disparities (Caucasian-African American) of each policy at final episode in realistic experiments.}\label{tab:real_rmse}\centering
\begin{tabular}{lrrrr}
\toprule
Policy Name & Mean $rMSE$ & SEM $rMSE$ & Mean $\Delta rMSE$ & SEM $\Delta rMSE$ \\
\midrule
Dis ($\tilde{D}=10$) & 0.0982 & 0.0008 & -0.0037 & 0.0016 \\
Dis ($\tilde{D}=20$) & 0.0845 & 0.0009 & -0.0094 & 0.0016 \\
Dis ($\tilde{D}=30$) & 0.0769 & 0.001 & -0.0059 & 0.0012 \\
Dis* ($\tilde{D}_1=10$, $c_R=125$) & 0.0845 & 0.0009 & -0.0043 & 0.0014 \\
Dis* ($\tilde{D}_1=10$, $c_R=5$) & 0.0978 & 0.0006 & -0.01 & 0.0016 \\
Dis* ($\tilde{D}_1=20$, $c_R=125$) & 0.0817 & 0.0009 & -0.0069 & 0.0013 \\
Dis* ($\tilde{D}_1=20$, $c_R=5$) & 0.095 & 0.0007 & -0.0095 & 0.0015 \\
Dis* ($\tilde{D}_1=30$, $c_R=125$) & 0.0813 & 0.001 & -0.0063 & 0.0017 \\
Dis* ($\tilde{D}_1=30$, $c_R=5$) & 0.0916 & 0.0007 & -0.0068 & 0.0016 \\
MaxLOS ($J=8$) & 0.1828 & 0.0105 & -0.0117 & 0.002 \\
State ($c_R=125$) & 0.1623 & 0.0096 & -0.0087 & 0.0022 \\
State ($c_R=40$) & 0.131 & 0.0068 & -0.009 & 0.002 \\
State ($c_R=5$) & 0.123 & 0.0011 & -0.0088 & 0.0018 \\
Thresh ($\tau=0.1$) & 0.1292 & 0.0021 & -0.0096 & 0.0019 \\
Thresh ($\tau=0.2$) & 0.1259 & 0.0012 & -0.0082 & 0.0015 \\
Thresh ($\tau=0.3$) & 0.1212 & 0.0009 & -0.0076 & 0.0016 \\
\bottomrule
\end{tabular}
\end{table}

\subsection{Extensions}\label{app:simulations:extensions}

\begin{figure}[ht]
\centering
\includegraphics[]{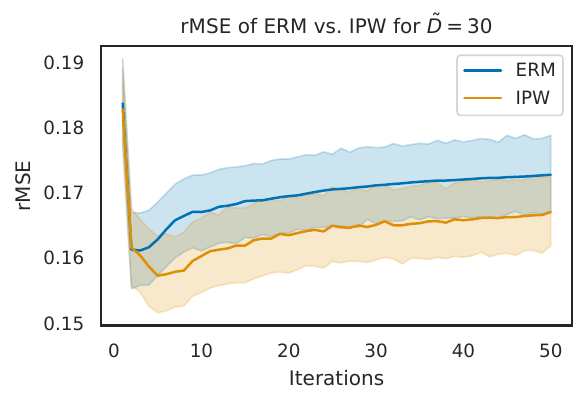}
\caption{ERM vs. IPW rMSE for the discharge-up-to level policy in the misspecified setting.}\label{fig:ipw}
\end{figure}

\paragraph{Misspecification.} We simulate all the policies in our simple setting with an inverse propensity weighting learning algorithm  where true risk is $r(\ell)=\sigma(b_0+b_1 \ell+b_2 \ell^2)$ such that $r(1)=0.7, r(8)=0.3, b_2=0.1$ and risk is misspecified as $\hat{r}(\ell)=\sigma(b_0+b_1 \ell)$ starting at $\hat{r}_1(1)=0.8$ and $\hat{r}_1(8)=0.2$. At the end of each episode, we simulate the discharge policy for $1000$ days to estimate the probability of observing each LOS. We then set weights on training data as the inverse queue probabilities of observing those discharge days, clipping weights to 20. Comparisons of ERM to IPW for the discharge-up-to policy $\tilde{D}=30$ are visualized in Figure \ref{fig:ipw}; we find IPW has a small advantage over ERM.

\begin{table}\small
\caption{Sample means and SEs for $c_R$ and $c_R$ disparities (Caucasian-African American) of each policy at final episode in realistic experiments.}\label{tab:real_cR}
\centering
\begin{tabular}{lrrrr}
\toprule
Policy Name & Mean Cost ($cR=5$) & SEM Cost ($cR=5$) & Mean Cost ($cR=125$) & SEM Cost ($cR=125$) \\
\midrule
Dis ($\tilde{D}=10$) & 541.6881 & 6.0977 & $1.33 \times 10^{4}$ & 146.7497 \\
Dis ($\tilde{D}=20$) & 832.7707 & 7.7814 & $1.18 \times 10^{4}$ & 131.5977 \\
Dis ($\tilde{D}=30$) & 1344.1148 & 8.3699 & $1.11 \times 10^{4}$ & 138.142 \\
Dis* ($\tilde{D}_1=10$, $c_R=125$) & NA & NA & $1.15 \times 10^{4}$ & 165.1165 \\
Dis* ($\tilde{D}_1=10$, $c_R=5$) & 555.0492 & 6.1148 & NA & NA \\
Dis* ($\tilde{D}_1=20$, $c_R=125$) & NA & NA & $1.12 \times 10^{4}$ & 183.4523 \\
Dis* ($\tilde{D}_1=20$, $c_R=5$) & 557.9974 & 7.59 & NA & NA \\
Dis* ($\tilde{D}_1=30$, $c_R=125$) & NA & NA & $1.16 \times 10^{4}$ & 180.9718 \\
Dis* ($\tilde{D}_1=30$, $c_R=5$) & 560.547 & 7.7171 & NA & NA \\
MaxLOS ($J=8$) & 2158.1319 & 20.4761 & $1.07 \times 10^{4}$ & 115.1221 \\
State ($c_R=125$) & NA & NA & $1.07 \times 10^{4}$ & 118.0508 \\
State ($c_R=5$) & 1212.6527 & 18.9713 & NA & NA \\
Thresh ($\tau=0.1$) & 1845.5076 & 23.017 & $1.06 \times 10^{4}$ & 132.0113 \\
Thresh ($\tau=0.2$) & 1229.1072 & 22.5611 & $1.10 \times 10^{4}$ & 108.2223 \\
Thresh ($\tau=0.3$) & 854.8178 & 15.7128 & $1.23 \times 10^{4}$ & 118.1589 \\
\bottomrule
\end{tabular}
\end{table}

\begin{table}\small
\caption{Sample means and SEs for $c_B$ and $c_B$ disparities (Caucasian-African American) of each policy at final episode in realistic experiments.}\label{tab:real_cB}
\centering
\begin{tabular}{lrrrr}
\toprule
Policy Name & Mean $\Delta c_R$ ($cR=5$) & SEM $\Delta c_R$ ($cR=5$) & Mean $\Delta c_R$ ($cR=125$) & SEM $\Delta c_R$ ($cR=125$) \\
\midrule
Dis ($\tilde{D}=10$) & -0.1976 & 0.0332 & -4.9409 & 0.8305 \\
Dis ($\tilde{D}=20$) & -0.2584 & 0.0353 & -6.4595 & 0.8824 \\
Dis ($\tilde{D}=30$) & -0.1809 & 0.0325 & -4.5226 & 0.8119 \\
Dis* ($\tilde{D}_1=10$, $c_R=125$) & NA & NA & -3.8986 & 0.852 \\
Dis* ($\tilde{D}_1=10$, $c_R=5$) & -0.2309 & 0.0443 & NA & NA \\
Dis* ($\tilde{D}_1=20$, $c_R=125$) & NA & NA & -4.4796 & 1.033 \\
Dis* ($\tilde{D}_1=20$, $c_R=5$) & -0.1929 & 0.0331 & NA & NA \\
Dis* ($\tilde{D}_1=30$, $c_R=125$) & NA & NA & -5.1936 & 0.9026 \\
Dis* ($\tilde{D}_1=30$, $c_R=5$) & -0.2701 & 0.031 & NA & NA \\
MaxLOS ($J=8$) & -0.1933 & 0.0373 & -4.8317 & 0.9319 \\
State ($c_R=125$) & NA & NA & -5.5815 & 0.9779 \\
State ($c_R=5$) & -0.1067 & 0.0359 & NA & NA \\
Thresh ($\tau=0.1$) & -0.1761 & 0.0356 & -4.403 & 0.8898 \\
Thresh ($\tau=0.2$) & -0.162 & 0.0356 & -4.0491 & 0.8905 \\
Thresh ($\tau=0.3$) & -0.1622 & 0.0283 & -4.0555 & 0.7079 \\
\bottomrule
\end{tabular}
\end{table}

\begin{table}
\caption{Sample means and SEs for $c_B$ and $c_B$ disparities (Caucasian-African American) of each policy at final episode in realistic experiments.}

\begin{tabular}{lrrrr}
\toprule
Policy Name & Mean $\Delta c_B$ ($cR=5$) & SEM $\Delta c_B$ ($cR=5$) & Mean $\Delta c_B$ ($cR=125$) & SEM $\Delta c_B$ ($cR=125$) \\
\midrule
Dis ($\tilde{D}=10$) & 0.0642 & 0.0078 & 1.6053 & 0.196 \\
Dis ($\tilde{D}=20$) & 0.1278 & 0.0093 & 3.1949 & 0.2328 \\
Dis ($\tilde{D}=30$) & 0.1513 & 0.01 & 3.7824 & 0.2498 \\
Dis* ($\tilde{D}_1=10$, $c_R=125$) & NA & NA & 2.7694 & 0.3119 \\
Dis* ($\tilde{D}_1=10$, $c_R=5$) & 0.0409 & 0.0086 & NA & NA \\
Dis* ($\tilde{D}_1=20$, $c_R=125$) & NA & NA & 2.893 & 0.2753 \\
Dis* ($\tilde{D}_1=20$, $c_R=5$) & 0.0356 & 0.0104 & NA & NA \\
Dis* ($\tilde{D}_1=30$, $c_R=125$) & NA & NA & 2.7714 & 0.282 \\
Dis* ($\tilde{D}_1=30$, $c_R=5$) & 0.0202 & 0.0087 & NA & NA \\
MaxLOS ($J=8$) & 0 & 0 & 0 & 0 \\
State ($c_R=125$) & NA & NA & 0 & 0 \\
State ($c_R=5$) & 0.144 & 0.0082 & NA & NA \\
Thresh ($\tau=0.1$) & 0.0858 & 0.0048 & 2.1453 & 0.1201 \\
Thresh ($\tau=0.2$) & 0.1393 & 0.0084 & 3.4829 & 0.2111 \\
Thresh ($\tau=0.3$) & 0.0848 & 0.0097 & 2.1196 & 0.2416 \\
\bottomrule
\end{tabular}
\end{table}

\begin{table}
\caption{Full significance test results for percent improvements.}\label{tab:significance}
\centering
\begin{tabular}{lrrrrr}
\toprule
Policy 1 & Policy 2 & Metric & \% & Decrease/Increase & $p$-value\\
\midrule
Dis* ($\tilde{D}_1=30$, $c_R=5$) & Thresh ($\tau=0.3$) & $rMSE$ & 33\% & Dec. & $1.33 \times 10^{-29}$\\
Dis* ($\tilde{D}_1=30$, $c_R=5$) & State ($c_R=5$) & $rMSE$ & 34\% & Dec. & $1.22 \times 10^{-29}$\\
Dis* ($\tilde{D}_1=30$, $c_R=5$) & Thresh ($\tau=0.3$) & $|\Delta rMSE|$ & 11\% & Dec. & 0.53 \\
Dis* ($\tilde{D}_1=30$, $c_R=5$) & State ($c_R=5$) & $|\Delta rMSE|$ & 23\% & Dec. & 0.47\\
Dis* ($\tilde{D}_1=20$, $c_R=5$) & Thresh ($\tau=0.3$) & Cost & 35\% & Dec. & $5.37\times 10^{-23}$\\
Dis* ($\tilde{D}_1=20$, $c_R=5$) & State ($c_R=5$) & Cost & 54\% & Dec. & $2.37\times 10^{-35}$\\
Dis* ($\tilde{D}_1=20$, $c_R=5$) & Thresh ($\tau=0.2$) & $|\Delta c_R|$ & 19\% & Inc. & $0.25$\\
Dis* ($\tilde{D}_1=20$, $c_R=5$) & State ($c_R=5$)  &  $|\Delta c_R|$ & 81\% & Inc.& $0.06$\\
Dis* ($\tilde{D}_1=20$, $c_R=5$) & Thresh ($\tau=0.3$) & $|\Delta c_B|$ & 57\% & Dec. & $0.0013$\\
Dis* ($\tilde{D}_1=20$, $c_R=5$) & State ($c_R=5$) & $|\Delta c_B|$ & 75\%  & Dec. & $1.39\times 10^{-12}$\\

\bottomrule
\end{tabular}
\end{table}

\end{document}